\pdfoutput=1
\documentclass[lettersize,journal]{IEEEtran}

\usepackage{amsmath,amsfonts}
\usepackage{array}
\usepackage[caption=false,font=normalsize,labelfont=sf,textfont=sf]{subfig}
\usepackage{textcomp}
\usepackage{stfloats}
\usepackage{placeins}
\usepackage{url}
\usepackage{verbatim}
\usepackage{graphicx}
\usepackage{cite}
\usepackage{booktabs}
\usepackage{multirow}
\usepackage{xcolor}

\newcommand{\qualcell}[1]{%
\begin{minipage}[c]{0.139\textwidth}%
\centering
\includegraphics[width=\linewidth]{#1}%
\end{minipage}}
\newcommand{\tempcell}[1]{\includegraphics[width=0.174\textwidth]{#1}}

\title{From Sparse and Imperfect 2D Anchors to Consistent 3D Gaussian Street Scenes: Support-Aware Appearance Distillation}

\author{Long Cao, Zhongquan Wang, Jie Li, Yuhan Chen, Kefei Qian, Xiangfei Huang, and Guofa Li%
\thanks{This work has been submitted to the IEEE for possible publication. This study is supported by the National Natural Science Foundation of China (Grant No. 52272421) and the State Key Laboratory of Intelligent Green Vehicle and Mobility under Project No. KFZ2409. (Corresponding author: Guofa Li.)}%
\thanks{Long Cao, Zhongquan Wang, Jie Li, Yuhan Chen, Kefei Qian, Xiangfei Huang, and Guofa Li are with the College of Mechanical and Vehicle Engineering, Chongqing University, Chongqing 400044, China. Email: caolong@stu.cqu.edu.cn; wangzhongquan@stu.cqu.edu.cn; jieli@cqu.edu.cn; 20240701028@stu.cqu.edu.cn; qiankf@stu.cqu.edu.cn; hxf0213@stu.cqu.edu.cn; liguofa@cqu.edu.cn.}}

\begin{document}

\maketitle

\begin{abstract}
Image priors can synthesize target conditions for 3D Gaussian street scenes, but independently edited views do not define a coherent 3D target. Direct fitting can propagate view-specific noise, while existing pipelines do not jointly handle imperfect sparse anchors and standard-rasterizer deployment. To address this gap, teacher-relative appearance residual distillation is introduced for appearance baking. A structured space for frequency decomposition, confidence estimation, and primitive-level lifting is formed by residuals between teacher anchors and original renders. The direct optimization signal is supplied by renderer-space matching, while primitive assignment is regularized by support-aware Gaussian-space aggregation. Supported detail is admitted and unsupported noise is suppressed through confidence-gated coarse-to-fine optimization, after which all residuals are baked into fixed-geometry spherical-harmonic coefficients. The teacher and auxiliary training modules are discarded at inference. Evaluation across Waymo street assets, Tanks and Temples scenes, and multiple target conditions shows a favorable overall balance of target alignment, content preservation, artifact suppression, and cross-view consistency over editing-based baselines. Ablations confirm the effectiveness of the main components. Code will be released at \url{https://github.com/Cagares/Baking-for-3D-Gaussian}.
\end{abstract}


\section{Introduction}

Controllable scenario generation supports simulation-based autonomous-driving testing \cite{lu2025omnitester}. Reconstructed static street assets within such simulations may need to adopt target visual conditions while supporting consistent novel-view rendering. Editing rendered frames independently is insufficient: the output should be a reusable 3D representation whose views share the same target appearance without propagating teacher-frame artifacts. 3D Gaussian Splatting (3DGS) is a suitable substrate because its explicit primitives support high-quality reconstruction and fast rasterization [1]. The focus is on static street-scene 3DGS assets that can be reused across camera views after a target appearance has been baked into the Gaussian representation.

The task is formulated as \emph{teacher-conditioned appearance baking} for a practical deployment setting: sparse teacher-defined appearance changes are baked into static 3DGS street assets as SH-only updates. An image-level teacher supplies anchors depicting conditions such as sunset, blue hour, or overcast dusk, and one target-specific 3DGS is optimized independently for each scene--condition pair. Gaussian positions, scales, rotations, and opacities remain fixed. After baking, the teacher and all auxiliary training modules are discarded, so deployment uses the unmodified 3DGS rasterizer with no additional inference-time modules.

Image relighting and editing models provide useful target-condition priors \cite{yang2025relighting,jin2024gaffer,zeng2024dilightnet,kim2024switchlight,phongthawee2024diffusionlight,chan2024anlightendiff,xing2025luminit,choi2025scribblelight,litman2025lightswitch,wang2025relighting}, while Gaussian editing methods propagate visual changes into 3D scenes \cite{liu2024stylegaussian,chen2026vip3de,lee2025editsplat}. The main difficulty is that supervision is provided only at a sparse set of camera views. Independently generated anchors may contain inconsistent illumination, boundary shifts, or hallucinated details, often appearing as noisy road or vegetation texture in street scenes. Directly fitting their RGB values can therefore overfit the sparse cameras and bake teacher-view artifacts into a scene that must also render unseen viewpoints. The key challenge is to distill a coherent, view-consistent appearance change from sparse and imperfect teacher anchors into shared Gaussian primitives.

To address this challenge, teacher-relative appearance residual distillation is introduced. The core innovation is support-aware residual coordination: noisy sparse 2D teacher evidence is organized into a confidence-aware residual space, lifted to shared Gaussian primitives, and baked as an SH-only 3DGS appearance update. Renderer-space residual matching provides the primary optimization signal for anchor-view appearance, while Gaussian-space aggregation acts as an auxiliary regularizer on how the change is distributed among shared primitives.

The baked SH appearance is further optimized in a confidence-gated coarse-to-fine manner. Low/mid-frequency components establish broad illumination and color changes, a constrained tone adapter refines global statistics, and an asymmetric high-frequency branch admits teacher-supported detail while suppressing unsupported detail injection. Because optimization is performed per target 3DGS rather than across scenes, procedural robustness is evaluated across street-scene assets, target conditions, anchor budgets, held-out views, and additional cross-dataset scenes.

The contributions are summarized as follows:
\begin{itemize}
    \item Teacher-conditioned appearance baking is formulated for static 3D Gaussian street-scene assets, where sparse teacher-defined appearance changes are baked as SH-only updates into one reusable 3DGS while preserving the standard inference pipeline.
    \item Support-aware residual coordination is introduced to convert noisy sparse 2D teacher evidence into a deployable SH-only 3DGS appearance update through confidence estimation, renderer-space matching, and auxiliary Gaussian-space aggregation.
    \item Confidence-gated coarse-to-fine SH appearance baking is developed to admit teacher-supported high-frequency changes, suppress unsupported detail injection, update only Gaussian appearance coefficients, and preserve the standard 3DGS inference path.
\end{itemize}

\section{Related Work}

\subsection{Gaussian Representations and View-Consistent Editing}

3D Gaussian Splatting (3DGS) \cite{kerbl3dgs} represents a scene with explicit primitives whose geometry, opacity, and spherical-harmonic (SH) appearance coefficients are rendered by differentiable rasterization. Subsequent work improves scale robustness, geometric fidelity, and large-scale urban reconstruction \cite{yu2024mipsplatting,lu2024scaffold,huang20242dgs,guedon2024sugar,lin2024vastgaussian,liu2024citygaussian,yan2024streetgaussians,zhou2024drivinggaussian}, making Gaussian scenes an increasingly practical representation for editable street scenes. On this substrate, StyleGaussian \cite{liu2024stylegaussian} embeds scene features into Gaussian primitives and transforms them according to a reference style image, followed by a 3D decoder for stylized rendering. ViP3DE \cite{chen2026vip3de} leverages a pretrained video-generation prior to synthesize consistent edited views from an edited conditioning view and uses them to update the underlying 3D representation. EditSplat \cite{lee2025editsplat} combines multi-view fusion guidance with attention-guided trimming for view-consistent text-driven editing. Together, they illustrate representative routes to general-purpose Gaussian appearance editing: reference-based stylization, video-prior propagation, and multi-view diffusion editing. These methods address general-purpose stylization or editing, rather than the specific problem of consolidating sparse, independently generated condition anchors that may disagree at the pixel level.

\subsection{Image-Conditioned Appearance Priors}

Image-level relighting, illumination editing, and enhancement methods can synthesize target conditions that are difficult to capture directly in a 3D street-scene sequence \cite{yang2025relighting,jin2024gaffer,zeng2024dilightnet,kim2024switchlight,phongthawee2024diffusionlight,chan2024anlightendiff,xing2025luminit,choi2025scribblelight,litman2025lightswitch,wang2025relighting,liu2023shadow,ma2023retinex,ye2024glow,wang2024pmsnet,luo2025unsupervised}. Such methods provide useful appearance teachers for generating target-condition anchors, including the GPT Image-2 API adopted in the experiments \cite{openai2026gptimage}. The mismatch arises when these image-space predictions are used as supervision for a shared 3D representation. Even when each image is plausible, independently generated anchors may disagree in illumination strength, shadow placement, color cast, object boundaries, or local detail. Directly fitting those RGB outputs would transfer view-specific teacher errors into the 3D representation rather than resolving them across shared primitives.

\subsection{Physical and Controllable 3D Relighting}

Physical and controllable relighting instead seeks explicit scene decomposition and lighting control. NeRF-based and multi-view inverse-rendering methods recover geometry, reflectance, visibility, and lighting for relightable view synthesis \cite{jin2023tensoir,liu2023nero,zhang2023neilf,schmitt2023scalable}. GIR \cite{shi2023gir}, included as a baseline, performs Gaussian-based inverse rendering by factorizing scene geometry, material properties, and illumination. Representative Gaussian relighting methods improve material--lighting decomposition and extend inverse rendering to reflective, outdoor, or sparsely observed scenes \cite{liang2024gsir,jiang2024gaussianshader,bi2024gs3,chen2025gigs,sun2025svgir,ye2025geosplatting,zhou2025rtrgs,liang2025gusir,lai2025glossygs,feng2025outdoor,fan2025rng,li2025recap,du2025gsid,bai2025gare,zhang2025surgs,pu2025invgs}. These works are central when the goal is interpretable lighting control or material-light factorization. The objective here is instead to reproduce a teacher-defined visual condition specified only through example images. Such a condition may combine illumination, tone, and local appearance changes that do not correspond to a calibrated environment map or a uniquely identifiable physical decomposition.

The proposed setting is positioned between these three lines, but is not interchangeable with any single one of them. StyleGaussian mainly transfers a reference style, ViP3DE propagates edits through a video prior, EditSplat relies on edited multi-view observations for text-driven fusion, and GIR targets physically motivated relighting through scene decomposition. In contrast, sparse and independently generated teacher anchors must be consolidated, unsupported teacher artifacts must be rejected, and a target-specific 3DGS must be deployed without a feature decoder, video prior, diffusion model, lighting estimate, or material factorization at inference time. Accordingly, teacher-relative appearance residuals are used as a support-aware coordination space: renderer-space residual matching provides direct anchor-view supervision, Gaussian-space aggregation regularizes how the change is distributed among shared primitives, and the resulting residual is baked into SH coefficients for standard 3DGS rendering.

\section{Method}

\subsection{Design Rationale}

A reconstructed static street-scene 3D Gaussian Splatting (3DGS) asset is converted into a target-condition-specific representation. The design targets sparse teacher-view appearance baking: target-condition examples must be consolidated into a reusable fixed-geometry 3DGS while unsupported teacher artifacts are rejected. The teacher provides appearance examples rather than geometry, material, or lighting parameters, so Gaussian positions, scales, rotations, and opacities remain fixed; only spherical-harmonic (SH) appearance coefficients are updated. Given an original scene $\mathcal{G}^{0}$, the baked scene $\mathcal{G}^{*}$ is rendered as
\begin{equation}
    \mathbf{I}^{*}(v)=\mathcal{R}(\mathcal{G}^{*},v),
\end{equation}
where $v$ denotes a camera view and $\mathcal{R}(\cdot)$ is the differentiable Gaussian renderer. The goal is not to recover physical illumination, but to bake a teacher-defined visual condition into a standard 3DGS asset.

\begin{figure*}[t]
    \centering
    \includegraphics[width=0.95\textwidth]{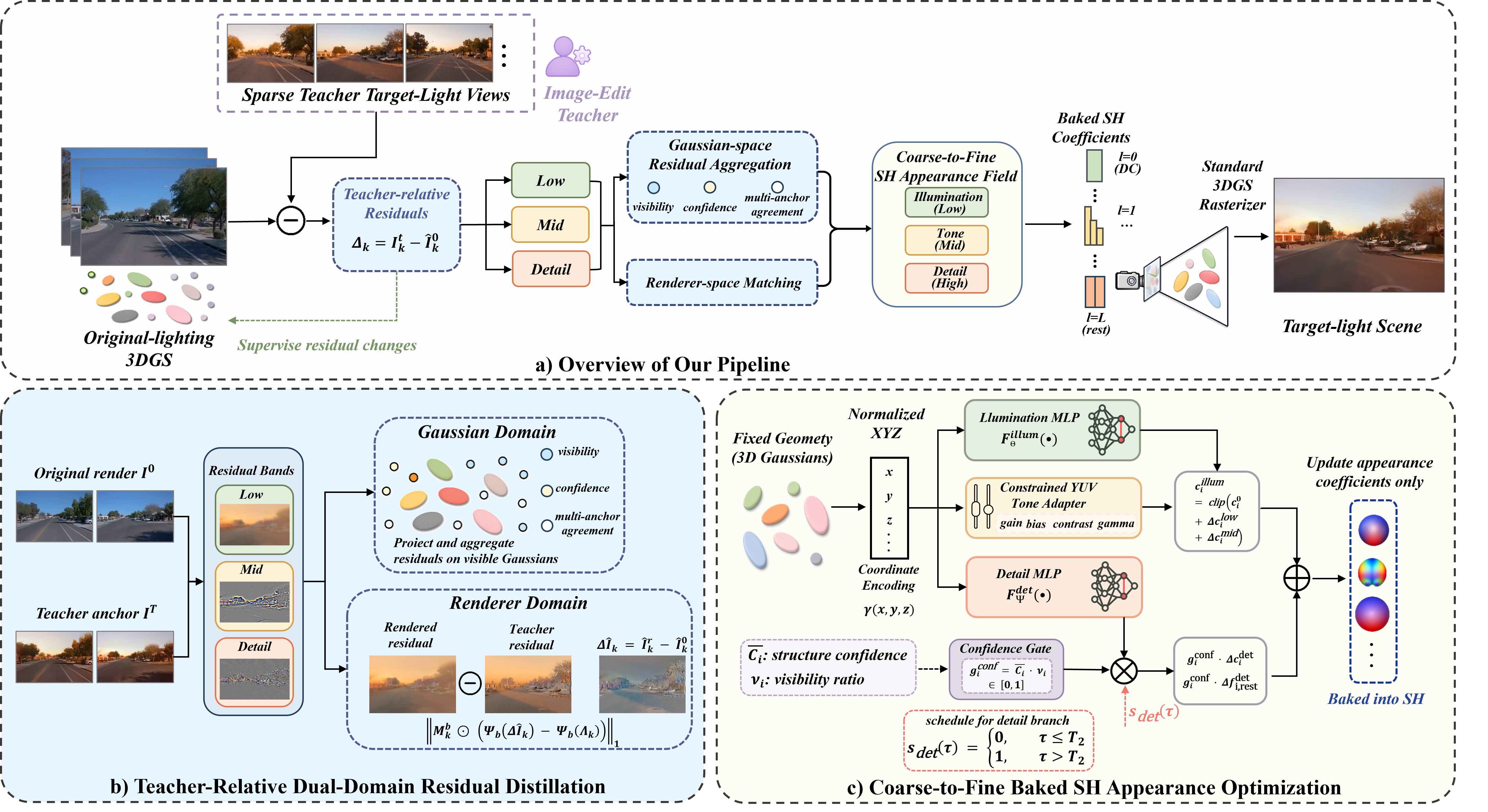}
    \caption{Overview of the framework. (a) Sparse teacher anchors define the target condition. (b) Teacher-relative residuals are distilled with support-aware artifact suppression. (c) Coarse-to-fine SH appearance baking produces the deployed 3DGS.}
    \label{fig:overview}
\end{figure*}

A direct teacher-fitting solution is fragile in this setting. Independently edited anchors may contain view-specific artifacts, inconsistent lighting, and noisy road or vegetation details. If these pixels are fitted directly, the optimized 3DGS can reproduce the anchor views while spreading teacher errors to non-anchor views. The method is therefore organized around four linked requirements. First, teacher evidence is represented as a change relative to the original render, which gives a common space for frequency decomposition and support estimation. Second, only structurally supported and repeatedly visible residuals are lifted to shared Gaussians. Third, high-frequency teacher details are treated asymmetrically: supported details are admitted, while unsupported injected details are suppressed. Finally, all temporary predictors are baked into SH coefficients, preserving the standard 3DGS inference path.

The two ablation groups in Sec.~IV follow this chain. The first group contrasts absolute teacher fitting, Gaussian-space-only lifting, renderer-space-only matching, and the full residual-baking model. The second group removes the Illumination MLP, Tone Adapter, Detail MLP, and progressive auxiliary decay. Thus, each reported variant corresponds to a specific failure mode rather than to an extra unreported component.

\subsection{Sparse Teacher Evidence as Residuals}

Let $\mathcal{A}=\{(v_k,\mathbf{I}^{t}_{k})\}_{k=1}^{K}$ be the sparse anchor set, where $v_k$ is an anchor camera and $\mathbf{I}^{t}_{k}$ is the teacher-edited target-condition image. The teacher is queried only at these anchor views and is discarded after optimization.

The first problem is that the teacher images are not ground truth renderings of a coherent 3D scene. They are sparse, independently edited examples and may include view-specific noise. Their evidence is therefore expressed as a change from the original 3DGS render:
\begin{equation}
    \hat{\mathbf{I}}^{0}_{k}=\mathcal{R}(\mathcal{G}^{0},v_k),\qquad
    \mathbf{R}^{t}_{k}=\mathbf{I}^{t}_{k}-\hat{\mathbf{I}}^{0}_{k}.
\end{equation}
The residual should be understood as an organizing representation. It lets the method ask where the teacher proposes a change, whether that change is structurally supported, and how the same change should be assigned to shared primitives. It is not used to claim that low/mid-frequency photometric matching becomes a fundamentally different objective. The residual is decomposed into broad illumination, regional shading, and local-detail bands:
\begin{equation}
\begin{aligned}
    \Psi_{\rm low}(\mathbf{R})&=G_{\sigma_l}*\mathbf{R},\\
    \Psi_{\rm mid}(\mathbf{R})&=G_{\sigma_m}*\mathbf{R}-G_{\sigma_l}*\mathbf{R},\\
    \Psi_{\rm hf}(\mathbf{R})&=\mathbf{R}-G_{\sigma_m}*\mathbf{R},
\end{aligned}
\end{equation}
where $G_{\sigma}(\cdot)$ is a fixed Gaussian low-pass filter, $*$ denotes convolution, and $\sigma_l>\sigma_m$.

Since Gaussian filtering is linear and the base render is fixed, low/mid residual matching satisfies
\begin{equation}
    \Psi_b(\hat{\mathbf{I}}^{r}_{k}-\hat{\mathbf{I}}^{0}_{k})-
    \Psi_b(\mathbf{I}^{t}_{k}-\hat{\mathbf{I}}^{0}_{k})
    =\Psi_b(\hat{\mathbf{I}}^{r}_{k}-\mathbf{I}^{t}_{k}),
    \quad b\in\{\mathrm{low},\mathrm{mid}\}.
\end{equation}
Thus, the technical role of the residual domain is not low/mid algebraic novelty, but a shared support-aware space for frequency decomposition, confidence estimation, primitive-level aggregation, and high-frequency control. The ablation in Table~\ref{tab:ablation_dual_domain} follows this interpretation: renderer-space matching supplies the direct anchor-view signal, while Gaussian-space lifting regularizes how this change is assigned to shared primitives.

\subsection{Support-Aware Lifting to Shared Gaussians}

The second problem is that an image residual is not automatically a valid primitive update. A residual pixel on a road surface, tree boundary, or sky region can come from a real target-condition change or from an editing artifact. Residuals are therefore lifted only when supported by three evidence sources. Structural support measures whether teacher and base images preserve compatible edges while avoiding newly introduced unmatched boundaries. Visibility support keeps only Gaussians that actually contribute to the anchor rendering. Cross-anchor support downweights primitive residuals that disagree across independently edited views.

The structural confidence map is computed from three interpretable cues:
\begin{equation}
\begin{aligned}
    \mathbf{C}_{k}
    &=\mathbf{C}^{\rm chg}_{k}\odot\mathbf{C}^{\rm edge}_{k}\odot\mathbf{C}^{\rm sup}_{k},\\
    \mathbf{C}^{\rm chg}_{k}
    &=\mathcal{N}(|G_{\sigma_m}*\mathbf{I}^{t}_{k}
    -G_{\sigma_m}*\hat{\mathbf{I}}^{0}_{k}|),\\
    \mathbf{C}^{\rm edge}_{k}
    &=\kappa(\nabla\hat{\mathbf{I}}^{0}_{k},\nabla\mathbf{I}^{t}_{k}),\\
    \mathbf{C}^{\rm sup}_{k}
    &=1-\mathcal{N}(||\nabla\mathbf{I}^{t}_{k}|-|\nabla\hat{\mathbf{I}}^{0}_{k}||).
\end{aligned}
\end{equation}
Here $\mathcal{N}(\cdot)$ denotes per-view normalization and $\kappa(\cdot,\cdot)$ is gradient-direction similarity. The first cue finds where the teacher proposes a condition change, the second preserves base-supported structures, and the third penalizes unmatched teacher edges.

For each visible projection of Gaussian $i$ in anchor $k$, the low, mid, and detail residual bands are sampled and aggregated into primitive-level targets:
\begin{equation}
\begin{aligned}
    \mathbf{p}_{ik}&=\Pi_{v_k}(\mathbf{x}_i),\qquad
    \mathbf{s}^{b}_{ik}=S(\Psi_b(\mathbf{R}^{t}_{k}),\mathbf{p}_{ik}),\\
    \bar{\mathbf{t}}^{b}_{i}
    &=\frac{\sum_k m_{ik}a_{ik}\mathbf{s}^{b}_{ik}}
    {\sum_k m_{ik}a_{ik}+\epsilon},
    \quad b\in\{\mathrm{low},\mathrm{mid},\mathrm{det}\}.
\end{aligned}
\end{equation}
where $\Pi_{v_k}$ is camera projection, $S(\cdot)$ is bilinear sampling, and $m_{ik}$ indicates raster visibility. The aggregation coefficient $a_{ik}$ samples $\mathbf{C}_{k}$ at $\mathbf{p}_{ik}$ and incorporates repeated observation. The auxiliary primitive weight further penalizes disagreement among independently edited anchors:
\begin{equation}
    w_i=\nu_i\bar{\mathbf{C}}_i
    \exp\left(-\frac{\mathrm{Var}_k(\mathbf{s}^{\rm low}_{ik}
    +\mathbf{s}^{\rm mid}_{ik}+\mathbf{s}^{\rm det}_{ik})}{\tau_{\rm agr}}\right).
\end{equation}
Thus large weights require visibility, structural support, repeated observation, and cross-anchor agreement.

This lifting step is intentionally auxiliary. It regularizes how a renderer-visible change is distributed among shared primitives, but it is not strong enough to realize the target condition by itself. This design choice matches Table~\ref{tab:ablation_dual_domain}: Gaussian-space Only has low non-anchor deviation because it under-edits, whereas Renderer-space Only loses structural and perceptual anchor fidelity. The full model uses Gaussian-space targets as support-aware regularization around the renderer-space signal.

\subsection{Coarse-to-Fine SH Baking and Detail Control}

The third problem is that target appearance is multi-scale. A sunset or overcast condition requires smooth color and illumination shifts, but the teacher may also add high-frequency artifacts on roads, vegetation, and object boundaries. The baked appearance update is therefore split into three scene-specific training modules that map each Gaussian location to SH-compatible residuals. With Fourier-encoded Gaussian position $\mathbf{z}_i=\gamma(\mathbf{x}_i)$, the Illumination MLP predicts low/mid-frequency color residuals:
\begin{equation}
    (\Delta\mathbf{c}^{\rm low}_i,\Delta\mathbf{c}^{\rm mid}_i)
    =F_{\theta}^{\rm illum}(\mathbf{z}_i).
\end{equation}
These terms carry brightness, color temperature, and regional shading. The Tone Adapter then applies a constrained YUV transform to the illumination-updated color:
\begin{equation}
    \mathbf{c}^{\rm tone}_i=
    F_{\phi}^{\rm tone}(\mathrm{clip}(\mathbf{c}^{0}_i+
    \Delta\mathbf{c}^{\rm low}_i+\Delta\mathbf{c}^{\rm mid}_i)),
\end{equation}
where $\mathbf{c}^{0}_i=\mathrm{SH2RGB}(\mathbf{f}^{0}_{i,\rm dc})$. It models global color cast, contrast, gamma, and saturation with a low-dimensional transformation rather than an unconstrained image generator. Finally, the Detail MLP proposes local DC and higher-order SH corrections:
\begin{equation}
    (\Delta\mathbf{c}^{\rm det}_i,\Delta\mathbf{f}^{\rm det}_{i,\rm rest})
    =F_{\psi}^{\rm det}(\mathbf{z}_i).
\end{equation}

These modules are deliberately not equal-purpose generators. The Illumination MLP and Tone Adapter establish the stable target direction, while the Detail MLP is accepted only through confidence-gated high-frequency control. For a rendered edited anchor $\hat{\mathbf{I}}^{r}_{k}$, the model residual is
\begin{equation}
    \Delta\mathbf{I}^{r}_{k}=\hat{\mathbf{I}}^{r}_{k}-\hat{\mathbf{I}}^{0}_{k}.
\end{equation}
Low/mid renderer-space matching aligns $\Delta\mathbf{I}^{r}_{k}$ with the teacher residual bands and provides the primary optimization signal. The lifted Gaussian targets from the previous subsection act as a primitive-level auxiliary loss and are decayed after the early stage, which is the Progressive Auxiliary Decay component in Table~\ref{tab:ablation_coarse_to_fine}.

High-frequency supervision is asymmetric. A detail mask is built from teacher residual strength, teacher edge strength, and structural confidence. Inside the mask, the model is allowed to match teacher-supported high-frequency residuals; outside it, newly introduced high-frequency residuals are penalized against the base-side reference. This mechanism is more important than treating the Detail MLP as a generic capacity increase: it admits local details only where the teacher change is supported and suppresses unsupported noise injection elsewhere. The qualitative comparisons and the Detail MLP ablation evaluate this behavior.

Concretely, the detail mask for iteration $\tau$ is
\begin{equation}
    \mathbf{M}^{(\tau)}_{k}
    =Q_{\rho(\tau)}(
    \mathcal{N}(|\Psi_{\rm hf}(Y(\mathbf{R}^{t}_{k}))|)
    \odot\mathcal{N}(|\nabla Y(\mathbf{I}^{t}_{k})|)
    \odot\mathbf{C}_{k}),
\end{equation}
where $Y(\cdot)$ is luminance and $Q_{\rho}$ keeps locations above quantile $\rho$. The quantile is relaxed after the global field stabilizes, allowing more teacher-supported details to enter later training. The teacher-side and base-side high-frequency terms are
\begin{equation}
\begin{aligned}
    \mathcal{L}_{\rm HF\text{-}T}
    &=\sum_k\|\mathbf{M}^{(\tau)}_k
    \odot(\Psi_{\rm hf}(\Delta Y^{r}_{k})-
    \Psi_{\rm hf}(\Delta Y^{t}_{k}))\|_1,\\
    \mathcal{L}_{\rm HF\text{-}B}
    &=\sum_k\|(1-\mathbf{M}^{(\tau)}_k)
    \odot\Psi_{\rm hf}(\Delta Y^{r}_{k})\|_1,
\end{aligned}
\end{equation}
with $\Delta Y^{r}_{k}=Y(\hat{\mathbf{I}}^{r}_{k})-Y(\hat{\mathbf{I}}^{0}_{k})$ and $\Delta Y^{t}_{k}=Y(\mathbf{I}^{t}_{k})-Y(\hat{\mathbf{I}}^{0}_{k})$. The first term admits reliable teacher detail; the second term suppresses unsupported high-frequency injection.

Optimization follows a coarse-to-fine schedule. Early iterations learn the global target direction from renderer-space low/mid losses and tone statistics. Later iterations activate spatial illumination and then the detail branch while reducing the Gaussian-space auxiliary weight. This order avoids letting noisy high-frequency residuals dominate before the stable target condition is established.

After the three modules are evaluated, their residuals are written to SH coefficients:
\begin{equation}
\begin{aligned}
    \mathbf{c}^{*}_i&=\mathrm{clip}(\mathbf{c}^{\rm tone}_i+
    s_{\rm det}\Delta\mathbf{c}^{\rm det}_i),\\
    \mathbf{f}^{*}_{i,\rm dc}&=\mathrm{RGB2SH}(\mathbf{c}^{*}_i),\\
    \mathbf{f}^{*}_{i,\rm rest}&=\mathbf{f}^{0}_{i,\rm rest}
    +s_{\rm det}\Delta\mathbf{f}^{\rm det}_{i,\rm rest}.
\end{aligned}
\end{equation}
The scalar $s_{\rm det}$ is controlled by the coarse-to-fine schedule, so higher-order SH changes are delayed until the low/mid target direction is stable.

\subsection{Optimization and Deployment}

The full objective combines five groups: renderer-space low/mid residual matching, tone/style statistics, confidence-gated high-frequency acceptance and suppression, auxiliary Gaussian-space residual targets, and weak magnitude/direction regularization. The renderer-space term compares the rendered residual with teacher residual bands:
\begin{equation}
\begin{aligned}
    \mathcal{L}_{\rm render}
    ={}&\sum_k\|\Psi_{\rm low}(\Delta\mathbf{I}^{r}_{k})
    -\Psi_{\rm low}(\mathbf{R}^{t}_{k})\|_1\\
    &+\sum_k\|\Psi_{\rm mid}(\Delta\mathbf{I}^{r}_{k})
    -\Psi_{\rm mid}(\mathbf{R}^{t}_{k})\|_1 .
\end{aligned}
\end{equation}
The auxiliary Gaussian-space term compares predicted primitive residuals with the lifted targets:
\begin{equation}
\begin{aligned}
    \mathcal{L}_{\rm Gauss}
    =\sum_i w_i(&\|\Delta\mathbf{c}^{\rm low}_i-\bar{\mathbf{t}}^{\rm low}_i\|_1
    +\|\Delta\mathbf{c}^{\rm mid}_i-\bar{\mathbf{t}}^{\rm mid}_i\|_1\\
    &+\bar{\mathbf{C}}_i\|\Delta\mathbf{c}^{\rm det}_i-\bar{\mathbf{t}}^{\rm det}_i\|_1).
\end{aligned}
\end{equation}
The total objective is kept grouped by role:
\begin{equation}
    \mathcal{L}_{\rm total}=
    \mathcal{L}_{\rm render}
    +\mathcal{L}_{\rm tone}
    +\mathcal{L}_{\rm HF}
    +\lambda_{\mathrm{3D}}\mathcal{L}_{\rm Gauss}
    +\mathcal{L}_{\rm reg}.
\end{equation}
Here $\mathcal{L}_{\rm HF}=\mathcal{L}_{\rm HF\text{-}T}+\mathcal{L}_{\rm HF\text{-}B}$, $\mathcal{L}_{\rm tone}$ contains global and patch-level YUV statistics, and $\mathcal{L}_{\rm reg}$ penalizes excessive residual magnitude and unnecessary higher-order SH drift. This grouping mirrors the two ablation tables: the first table tests renderer-space versus Gaussian-space evidence, and the second table tests the coarse-to-fine appearance components.

The component scales follow
\begin{equation}
    (s_{\rm illum}(\tau),s_{\rm det}(\tau))=
    \begin{cases}
    (0,0), & \tau\le T_1,\\
    (1,0), & T_1<\tau\le T_2,\\
    (1,1), & \tau>T_2,
    \end{cases}
\end{equation}
and the Gaussian auxiliary weight is reduced after the early stage. This schedule gives the renderer-space signal priority, then uses Gaussian-space and detail constraints to regularize the final baked appearance.

After optimization, the temporary modules are evaluated once over all Gaussians and their predicted residuals are written into the SH appearance coefficients:
\begin{equation}
    \mathcal{G}^{*}=\mathcal{G}^{0}\oplus\Delta\mathcal{G}_{\rm SH},
\end{equation}
where $\Delta\mathcal{G}_{\rm SH}$ changes only SH appearance. Inference then uses the unmodified 3DGS rasterizer, $\mathbf{I}^{*}(v)=\mathcal{R}(\mathcal{G}^{*},v)$. No confidence map, agreement computation, residual predictor, lighting estimate, material decomposition, shadow model, diffusion model, or HDR environment map is evaluated at test time.

\section{Experiments}

\subsection{Experimental Setup}

The experiments are organized around target appearance, content preservation, asset-level stability, residual-domain design, and the renderer/Gaussian/coarse-to-fine components. Macro-averaged comparisons, scene-cluster bootstrap intervals, fixed qualitative cases, a consecutive-view sequence, and controlled ablations are used.

\subsubsection{Dataset and target setting}
Four static street-scene assets are selected from the Waymo Open Dataset \cite{sun2020waymo}, with numeric IDs 019, 1067, 1172, and 1776, three target conditions (sunset, blue hour, and overcast dusk), and 198 ordered camera views per scene--condition group. This gives $4\times3\times198=2376$ view--condition instances across 12 independently optimized baked 3DGS models. Each Base 3DGS is trained for 30k iterations with SH degree 3; one appearance is optimized per target condition under fixed geometry. For each group, 16 teacher anchors are selected approximately uniformly along the trajectory, giving 192 anchor images in total. Anchors supervise optimization and the remaining 182 views per group serve as non-anchor evaluation views. Train and Truck from Tanks and Temples \cite{knapitsch2017tanks} are additionally used with the same 16-anchor setting to evaluate the per-scene baking procedure on non-Waymo 3D assets.

\subsubsection{Baselines}
Four routes to the task are compared: StyleGaussian \cite{liu2024stylegaussian} for reference-image Gaussian style transfer, GIR \cite{shi2023gir} for inverse-rendering-based relighting, ViP3DE \cite{chen2026vip3de} for video-prior multi-view editing, and EditSplat \cite{lee2025editsplat} for text-guided 2D editing with multi-view fusion. For ViP3DE and EditSplat, the input side is restricted to the same sparse teacher anchors to form a controlled sparse-anchor comparison under the supervision budget targeted here, rather than a claim over their fully dense default pipelines. Default settings are retained for GIR and StyleGaussian: an environment map close to the target lighting is used by GIR, and one anchor is randomly selected as the reference for StyleGaussian. All methods are exported at the same 198 ordered camera identifiers and scored with the same prompts, Base renders, OpenCLIP checkpoint, resizing, adjacency rule, and metrics.

\subsubsection{Implementation details}
Teacher anchors are generated offline with the GPT Image-2 API \cite{openai2026gptimage}, which is not used at render time, and are not manually filtered; the setup therefore preserves realistic teacher noise such as road and vegetation artifacts. Fig.~\ref{fig:dual_domain_ablation} additionally uses a Qwen image-edit teacher as a qualitative teacher-dependence check. Each scene--condition optimization runs for 5k Adam iterations \cite{kingma2015adam} with predictor learning rate $10^{-3}$; hidden dimensions are 64 for illumination/tone and 32 for detail, with four Fourier bands. Gaussian filters $(k,\sigma)=(21,5.0)$ and $(7,1.5)$ define the residual bands. The 30k-iteration SH-degree-3 Base models, 16 anchors, and Full configuration are fixed for all 12 runs. Only scene-specific residual predictors are optimized and baked into SH coefficients. On one RTX 4090, optimization averages 5.5 minutes per scene--condition and renders at 168 fps. GIR is trained for 60k iterations; other baselines provide edited sequences at the common camera identifiers.

\subsubsection{Metrics}
Four compact diagnostics are used as complementary signals, with formulas deferred to the supplementary material. CLIP-S and CLIP-DS use CLIP-style image--text similarities \cite{radford2021clip} with OpenCLIP ViT-B/32 \texttt{laion2b\_s34b\_b79k} weights \cite{cherti2023openclip}. CLIP-S checks source-scene compatibility, CLIP-DS measures movement toward the target condition, Realism is a prompt-margin artifact diagnostic rather than a human perceptual score, and CV-$\Delta$ compares adjacent-view changes with the corresponding Base changes. The blind human study complements these automatic diagnostics by directly comparing perceived realism, content preservation, target-condition match, and cross-view consistency. RGB images are resized to $256\times256$. Main-benchmark scores are macro-averaged over 12 Waymo scene--condition groups, and Tanks and Temples scores are reported separately. CLIP-S, CLIP-DS, and Realism are higher-is-better; CV-$\Delta$ is lower-is-better.

\subsubsection{Statistical uncertainty}
The three conditions from the same asset are not treated as independent samples. A scene-cluster bootstrap is applied over the four assets, with all three conditions retained for each sampled asset and 10,000 resamples generated using random seed 20260614. Because each target 3DGS is optimized independently rather than through a cross-scene model, the intervals evaluate procedural robustness across assets, conditions, anchor budgets, and held-out views.

\begin{table}[t]
\centering
\caption{Quantitative comparison over 12 scene--condition groups.}
\label{tab:main}
\scriptsize
\setlength{\tabcolsep}{1.8pt}
\begin{tabular}{lcccc}
\toprule
Method & CLIP-S $\uparrow$ & CLIP-DS $\uparrow$ & Realism $\uparrow$ & CV-$\Delta$ $\downarrow$\\
\midrule
StyleGaussian & 0.8139 & 0.1349 & 0.0343 & 0.0319\\
GIR & 0.8282 & 0.1166 & 0.0137 & 0.0314\\
ViP3DE & 0.8435 & 0.2046 & 0.0501 & 0.0283\\
EditSplat & 0.8815 & 0.2121 & 0.0536 & 0.0231\\
Ours & \textbf{0.8980} & \textbf{0.2574} & \textbf{0.0764} & \textbf{0.0116}\\
\bottomrule
\end{tabular}
\end{table}

\begin{table}[t]
\centering
\caption{Sparse-anchor comparison on two Waymo scene--condition groups.}
\label{tab:sparse_anchor}
\scriptsize
\setlength{\tabcolsep}{1.8pt}
\begin{tabular}{@{}lcccc@{}}
\toprule
Method & Anchors & CLIP-S $\uparrow$ & CLIP-DS $\uparrow$ & CV-$\Delta$ $\downarrow$\\
\midrule
Ours & 4 & \textbf{0.9498} & \textbf{0.2243} & \textbf{0.0106}\\
Ours & 8 & \textbf{0.9548} & \textbf{0.2403} & \textbf{0.0107}\\
EditSplat & 4 & 0.7641 & 0.1482 & 0.0470\\
EditSplat & 8 & 0.8145 & 0.1547 & 0.0438\\
ViP3DE & 4 & 0.8008 & 0.1303 & 0.0629\\
ViP3DE & 8 & 0.7951 & 0.1539 & 0.0529\\
\bottomrule
\end{tabular}
\end{table}

\begin{table}[t]
\centering
\caption{Cross-dataset results on Tanks and Temples Train and Truck.}
\label{tab:cross_dataset}
\scriptsize
\setlength{\tabcolsep}{2.0pt}
\begin{tabular}{@{}lcccc@{}}
\toprule
Method & CLIP-S $\uparrow$ & CLIP-DS $\uparrow$ & Realism $\uparrow$ & CV-$\Delta$ $\downarrow$\\
\midrule
Ours & \textbf{0.8890} & \textbf{0.1209} & \textbf{0.6421} & \textbf{0.1286}\\
EditSplat & 0.8785 & 0.0284 & 0.6360 & 0.1524\\
ViP3DE & 0.5066 & 0.0624 & 0.6335 & 0.1505\\
\bottomrule
\end{tabular}
\end{table}

\begin{table}[t]
\centering
\caption{Randomized blinded perceptual study with 30 participants.}
\label{tab:human_study}
\scriptsize
\setlength{\tabcolsep}{3.0pt}
\begin{tabular}{@{}llc@{}}
\toprule
Group & Comparison & Decisive pref. $\uparrow$\\
\midrule
Overall & 731/70/129 & \textbf{85.0 [82.5, 87.2]}\\
\midrule
Dimension & Visual realism & 85.9\\
Dimension & Content preservation & 85.4\\
Dimension & Target alignment & 84.9\\
Dimension & Cross-view consistency & 82.2\\
\midrule
Baseline & vs. GIR & 93.2\\
Baseline & vs. StyleGaussian & 89.7\\
Baseline & vs. ViP3DE & 83.6\\
Baseline & vs. EditSplat & 70.3\\
\bottomrule
\end{tabular}
\end{table}

\subsection{Main Results}

Table~\ref{tab:main} reports the macro average over all 12 scene--condition groups. The proposed method ranks first among edited methods on all four reported metrics, showing the best overall balance between target realization and cross-view consistency.

Table~\ref{tab:sparse_anchor} evaluates the same uniformly selected 4 or 8 anchors on two Waymo cases. With four anchors, the proposed method reaches CLIP-DS 0.2243 and CV-$\Delta$ 0.0106, compared with 0.1482/0.0470 for EditSplat and 0.1303/0.0629 for ViP3DE. The same ranking is retained with eight anchors, while Fig.~\ref{fig:sparse_visual_main} shows less underfitting and scene degradation.

Table~\ref{tab:cross_dataset} reports results on Train and Truck. Compared with EditSplat, CLIP-DS increases by 0.0925 and CV-$\Delta$ decreases by 0.0238; CLIP-S and Realism also remain higher. These results indicate that the same per-scene appearance-baking procedure remains effective on non-Waymo 3D assets.

Table~\ref{tab:human_study} reports a randomized and blinded A/B perceptual study in which the proposed method was compared with one baseline at a time under matched scene, view, and condition. Four criteria and a tie option were presented to 30 participants. The proposed method was preferred in 731 judgments (78.6\%), with 70 ties and 129 baseline wins. After ties were excluded, an 85.0\% preference rate was obtained (95\% CI: 82.5--87.2\%); aggregate preferences favored the proposed method for all 30 participants (two-sided sign test, $p<0.001$).

\subsubsection{Metric-wise analysis}
The embedding metrics capture complementary behavior: CLIP-S measures source compatibility, CLIP-DS measures the requested feature displacement, and Realism is an artifact-prompt margin rather than a human judgement.

CV-$\Delta$ can favor conservative outputs because Base obtains zero without performing an edit. It is therefore interpreted only together with target alignment.

\begin{figure}[t]
\centering
\includegraphics[width=\columnwidth]{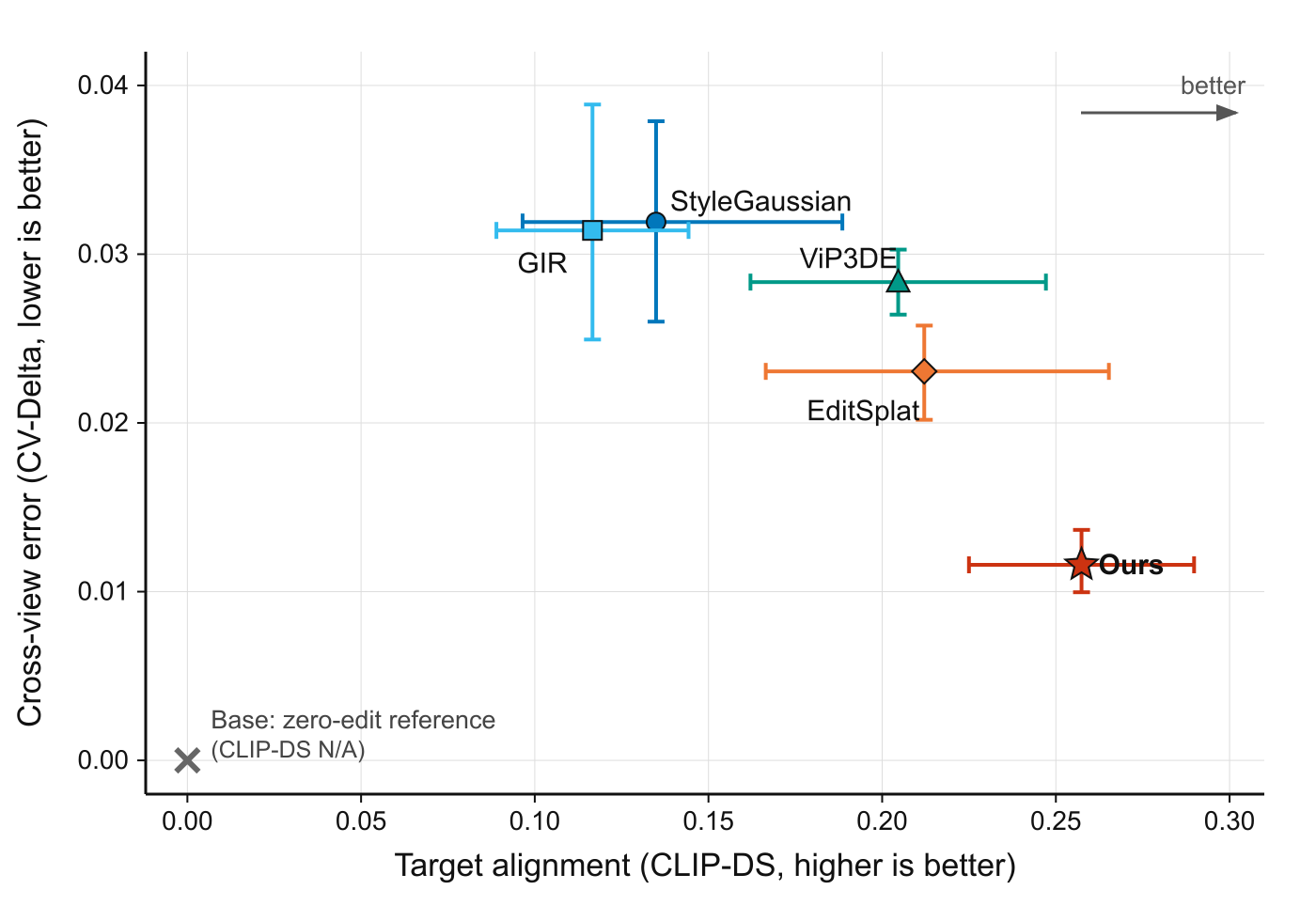}
\caption{Target alignment versus cross-view error with 95\% scene-cluster bootstrap intervals. Base is the zero-edit reference; Pareto probabilities are computed among edited methods.}
\label{fig:baseaware_pareto}
\end{figure}

Fig.~\ref{fig:baseaware_pareto} and Table~\ref{tab:baseaware_main} jointly evaluate alignment and consistency. Base is included only as a zero-edit endpoint: CV-$\Delta$ is zero by construction and CLIP-DS is undefined. Among edited methods, the proposed method is the only non-dominated result and has Pareto probability 1.000. Relative to EditSplat, CLIP-DS increases by 0.04509 [0.00965, 0.09371] and CV-$\Delta$ decreases by 0.01145 [0.00888, 0.01405]; both scene-cluster intervals exclude zero.

\begin{table}[t]
\centering
\caption{Base-aware target alignment and cross-view analysis with 95\% scene-cluster CIs.}
\label{tab:baseaware_main}
\scriptsize
\setlength{\tabcolsep}{1.4pt}
\begin{tabular}{@{}lccc@{}}
\toprule
Method & CLIP-DS $\uparrow$ & CV-$\Delta$ $\downarrow$ & Pareto\\
\midrule
Base & N/A & 0.00000 [0.00000, 0.00000] & Ref.\\
StyleGaussian & 0.1349 [0.0965, 0.1885] & 0.03190 [0.02600, 0.03788] & 0.000\\
GIR & 0.1166 [0.0890, 0.1443] & 0.03141 [0.02494, 0.03887] & 0.000\\
ViP3DE & 0.2046 [0.1621, 0.2471] & 0.02834 [0.02642, 0.03027] & 0.000\\
EditSplat & 0.2121 [0.1665, 0.2653] & 0.02305 [0.02018, 0.02577] & 0.000\\
Ours & \textbf{0.2574 [0.2250, 0.2898]} & \textbf{0.01159 [0.00996, 0.01366]} & \textbf{1.000}\\
\bottomrule
\end{tabular}
\end{table}

\subsection{Qualitative Results}

\begin{figure*}[t]
\centering
\scriptsize
\setlength{\tabcolsep}{0.8pt}
\renewcommand{\arraystretch}{1.02}
\begin{tabular}{@{}p{0.065\textwidth}cccccc@{}}
& \textbf{Teacher} & \textbf{StyleGaussian} & \textbf{GIR} & \textbf{ViP3DE} & \textbf{EditSplat} & \textbf{Ours}\\
\multicolumn{7}{@{}l}{\textbf{Waymo scene A}}\\[-0.2ex]
Sunset &
\qualcell{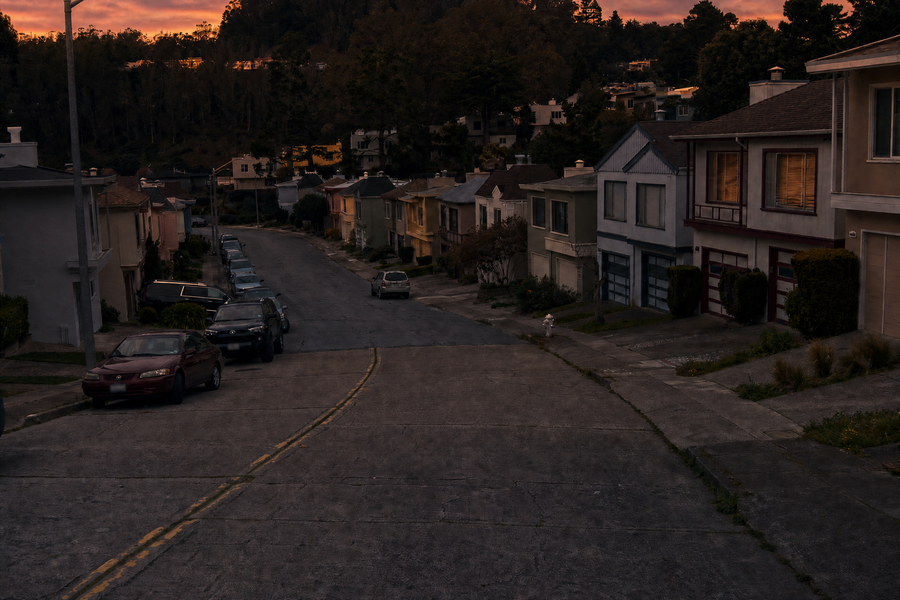} &
\qualcell{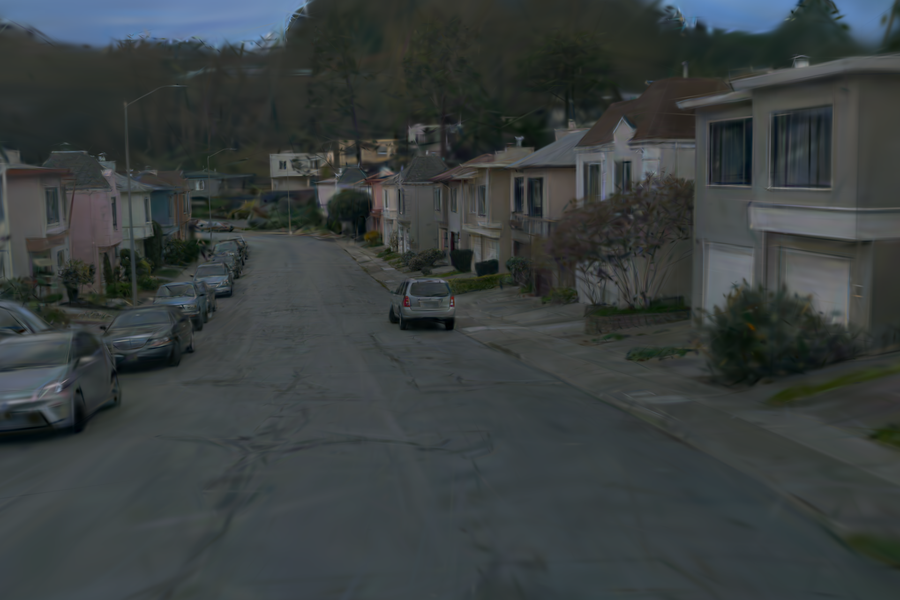} &
\qualcell{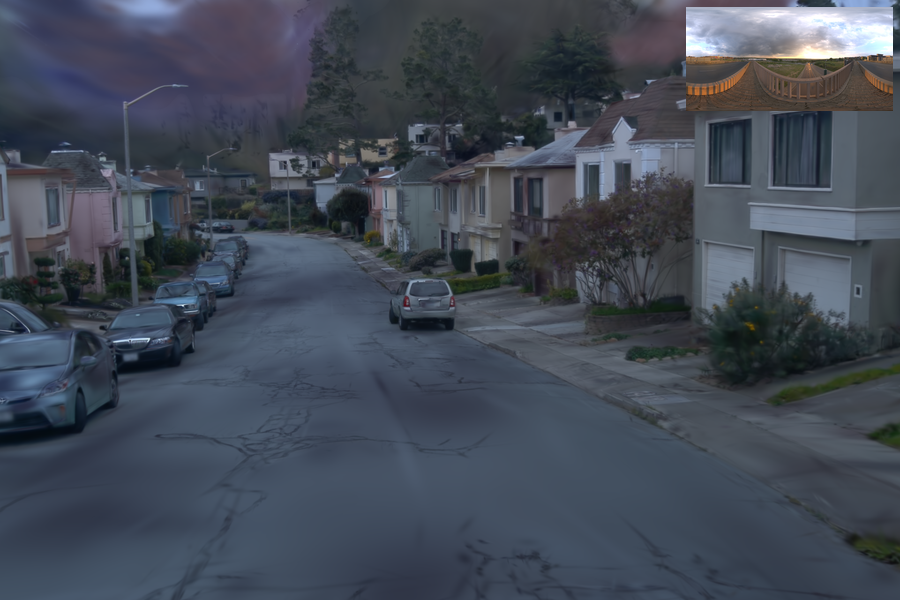} &
\qualcell{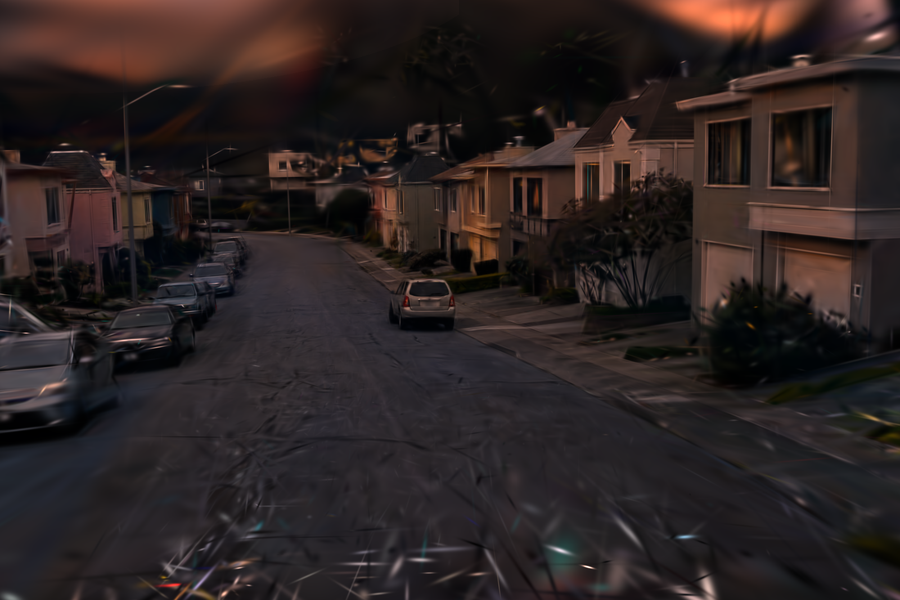} &
\qualcell{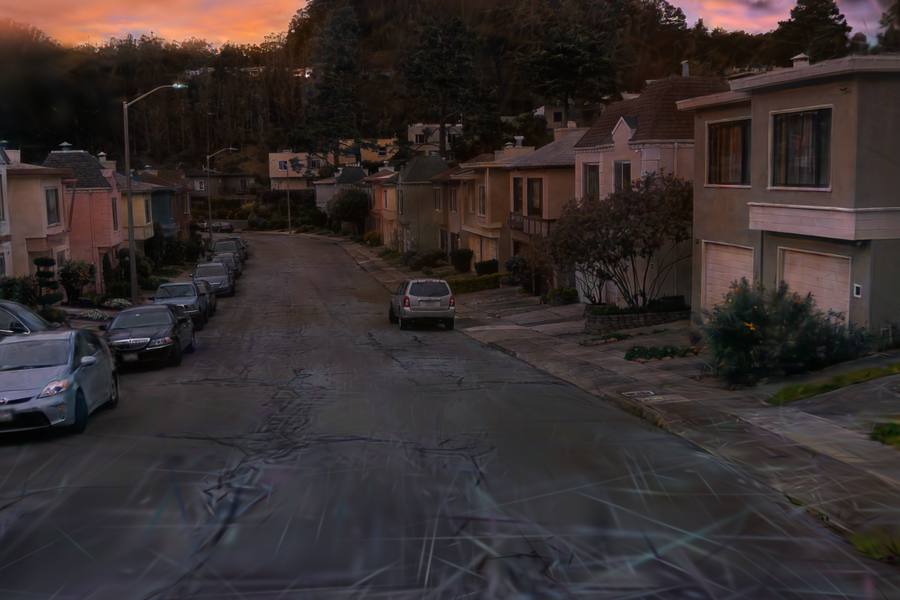} &
\qualcell{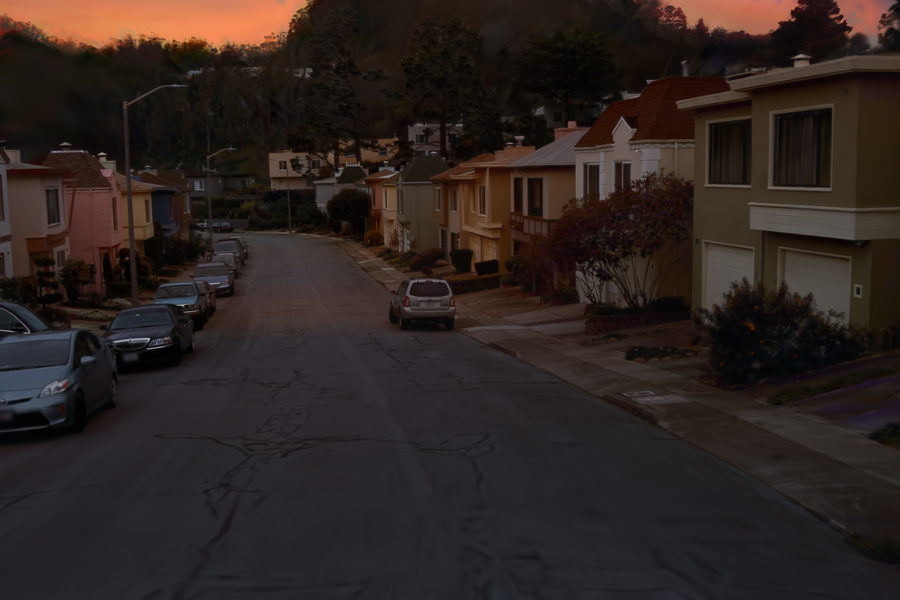}\\[0.3ex]
Blue hour &
\qualcell{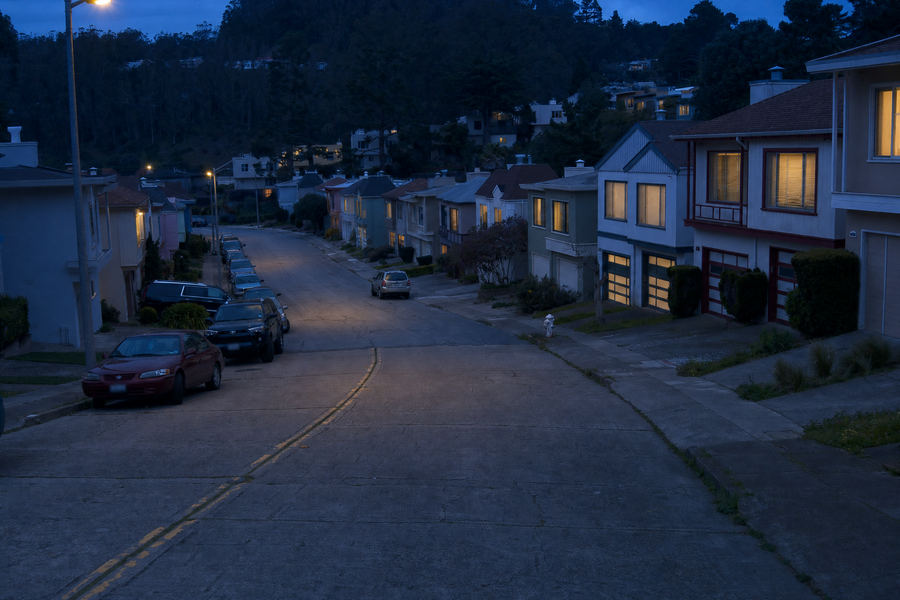} &
\qualcell{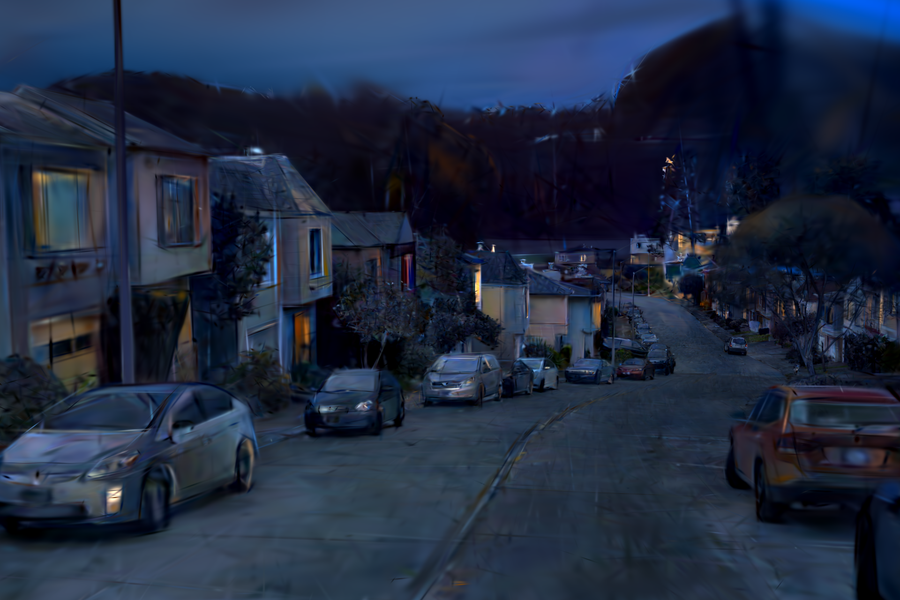} &
\qualcell{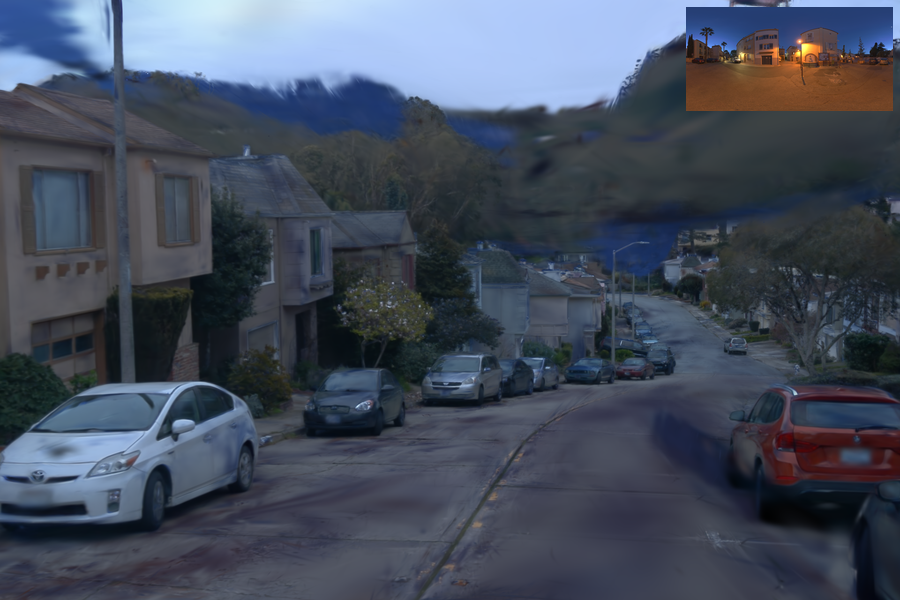} &
\qualcell{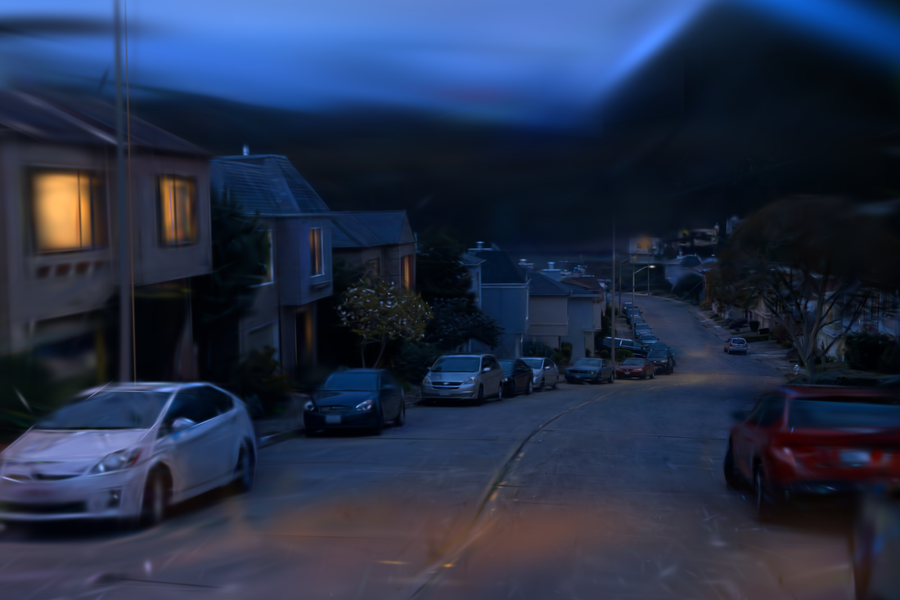} &
\qualcell{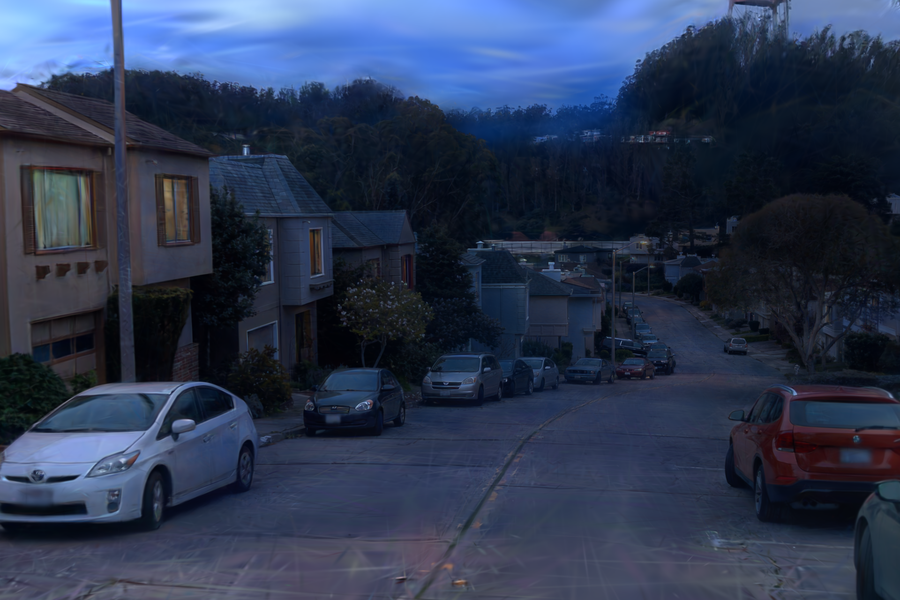} &
\qualcell{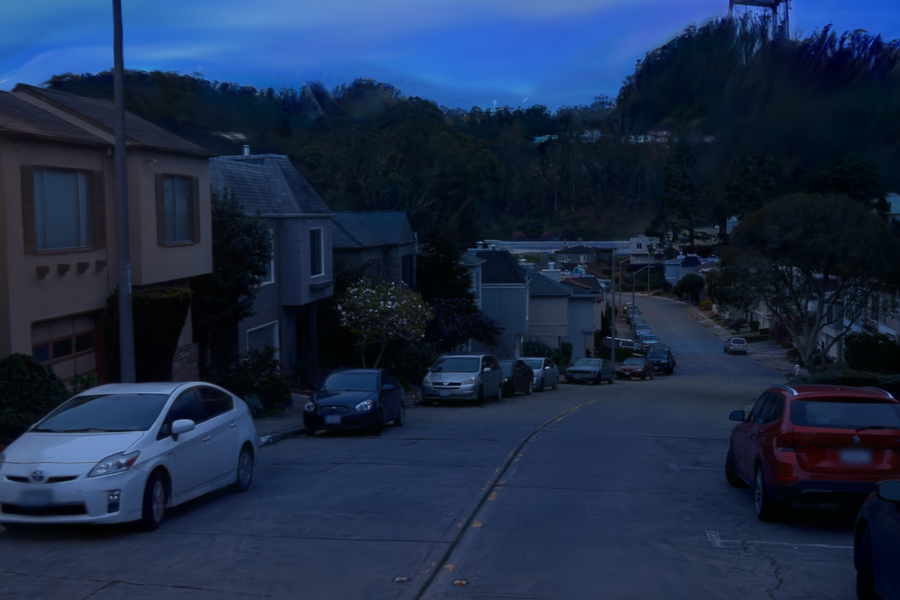}\\[0.3ex]
Overcast &
\qualcell{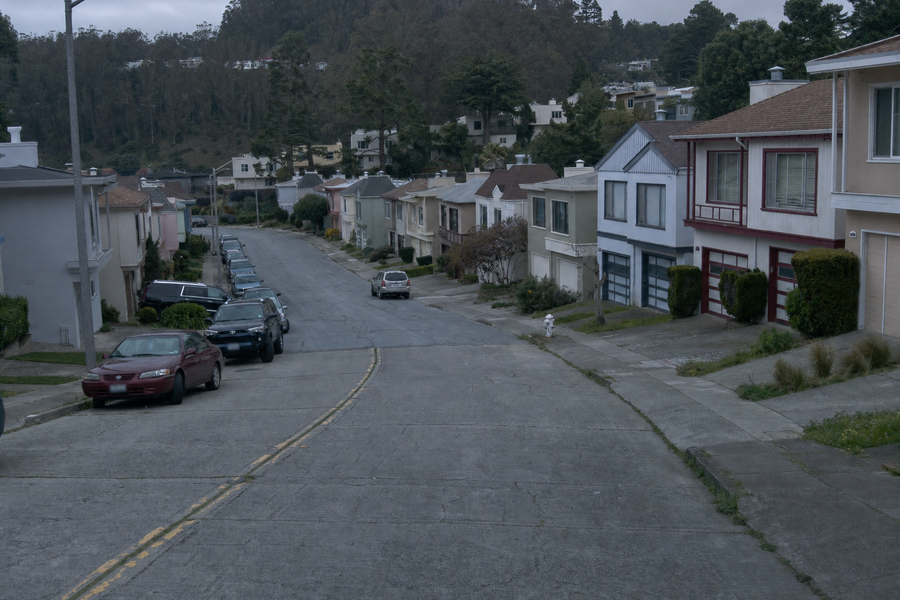} &
\qualcell{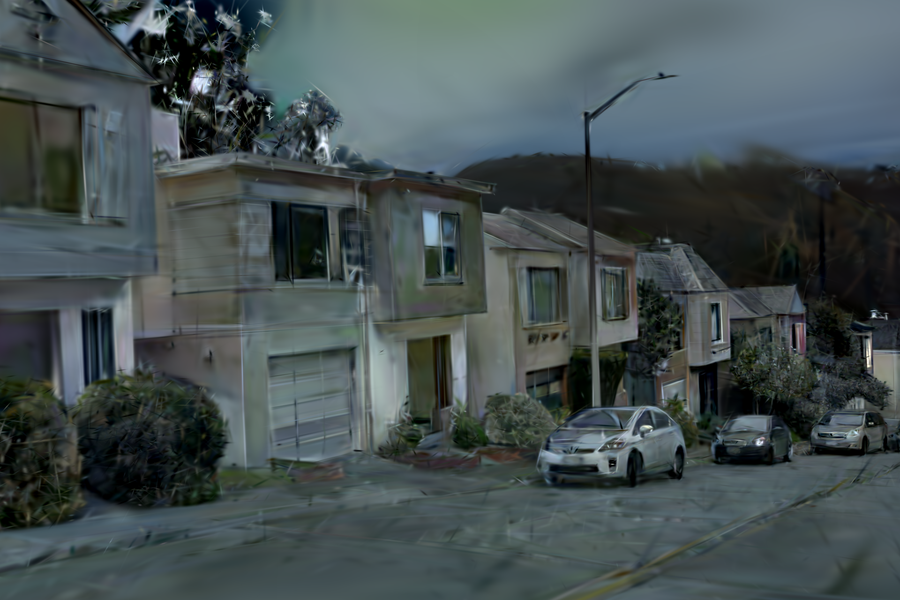} &
\qualcell{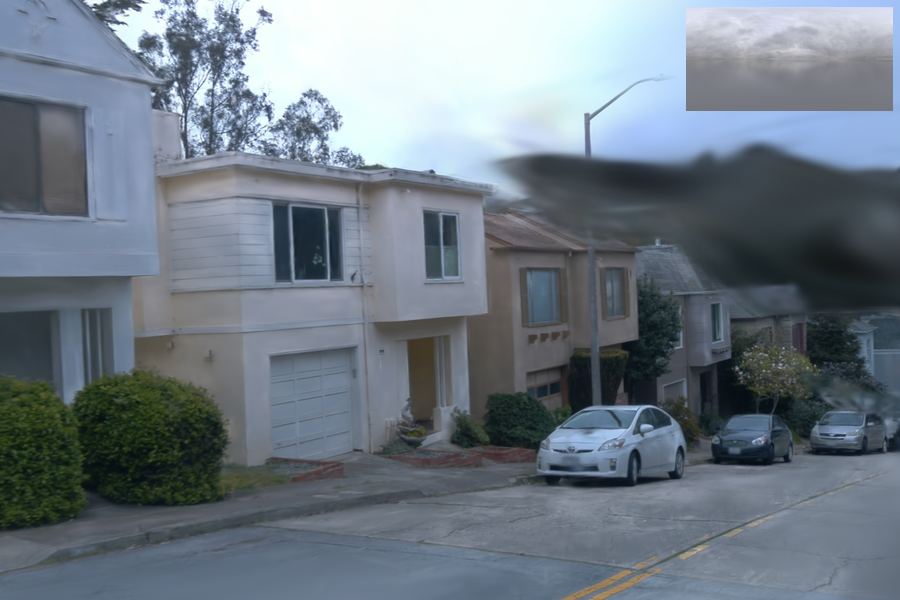} &
\qualcell{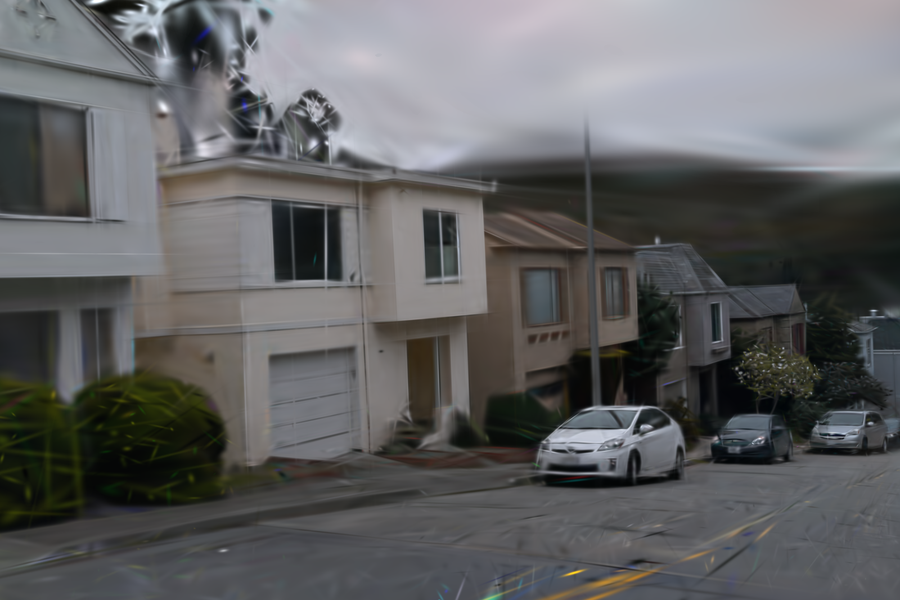} &
\qualcell{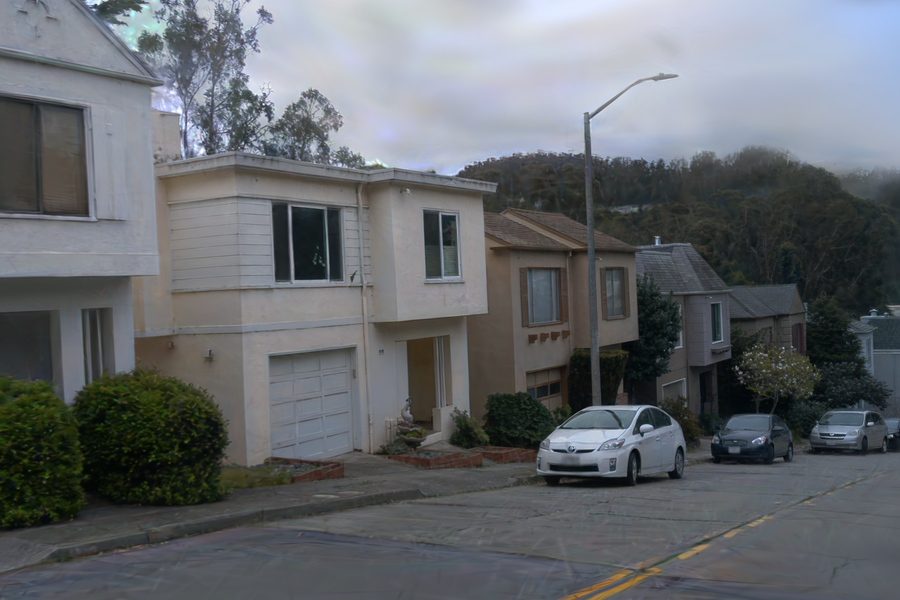} &
\qualcell{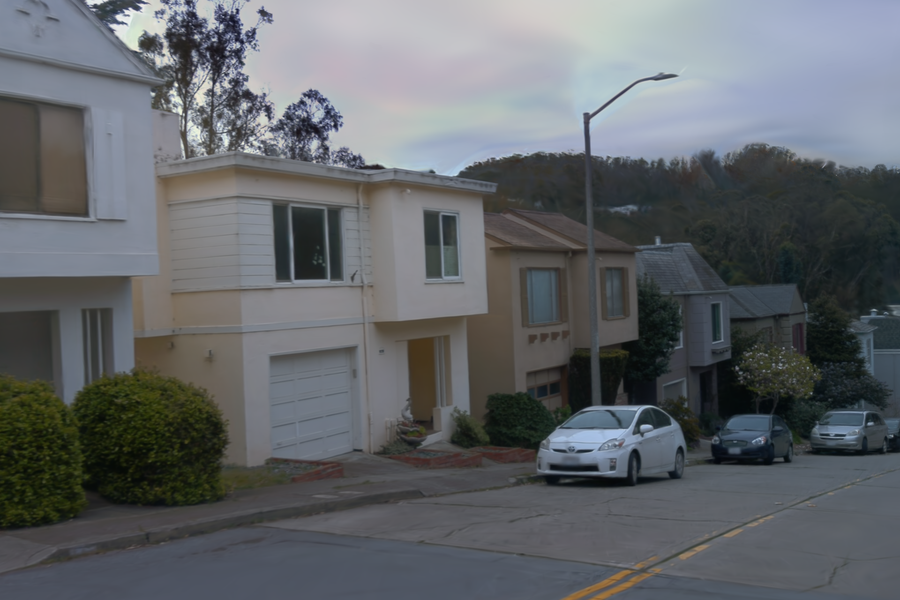}\\[0.8ex]
\multicolumn{7}{@{}l}{\textbf{Waymo scene B}}\\[-0.2ex]
Sunset &
\qualcell{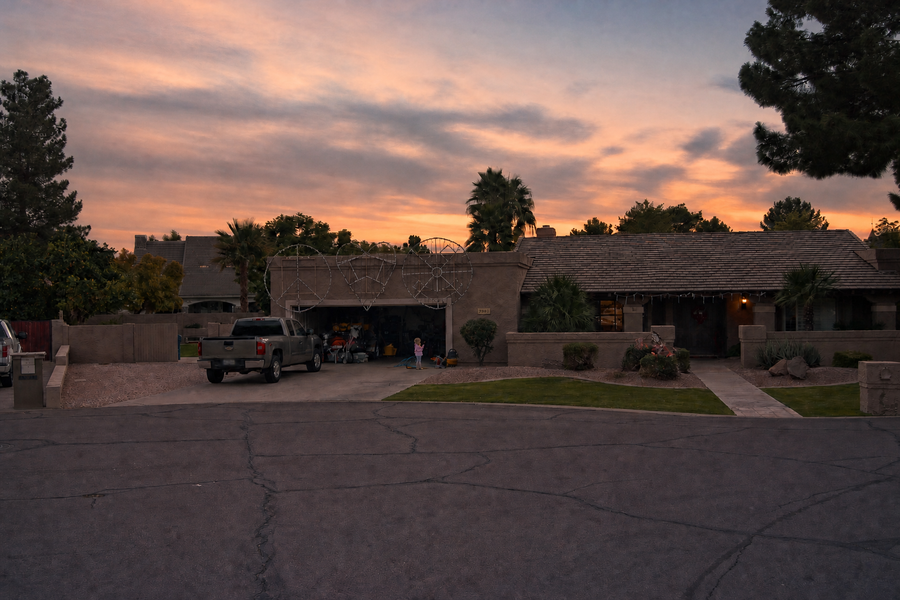} &
\qualcell{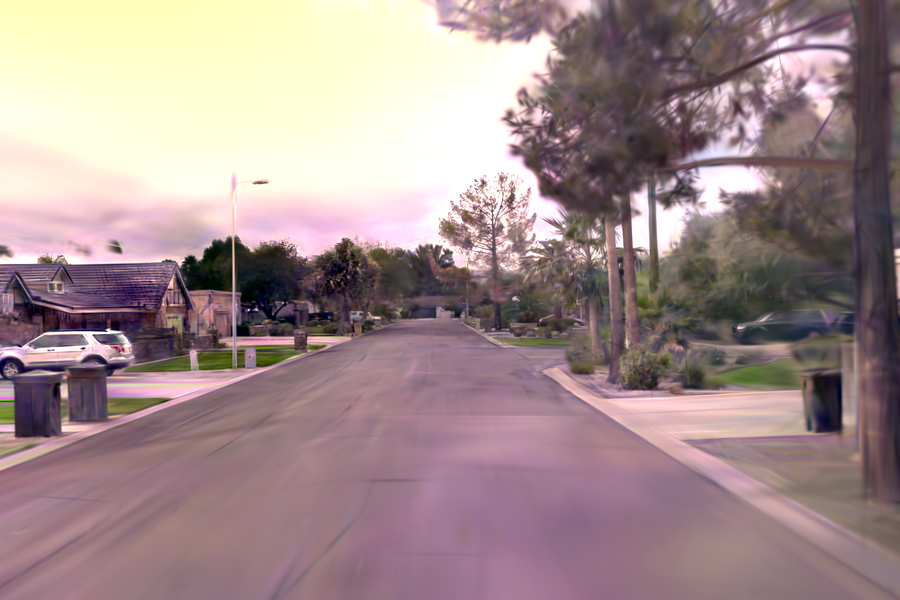} &
\qualcell{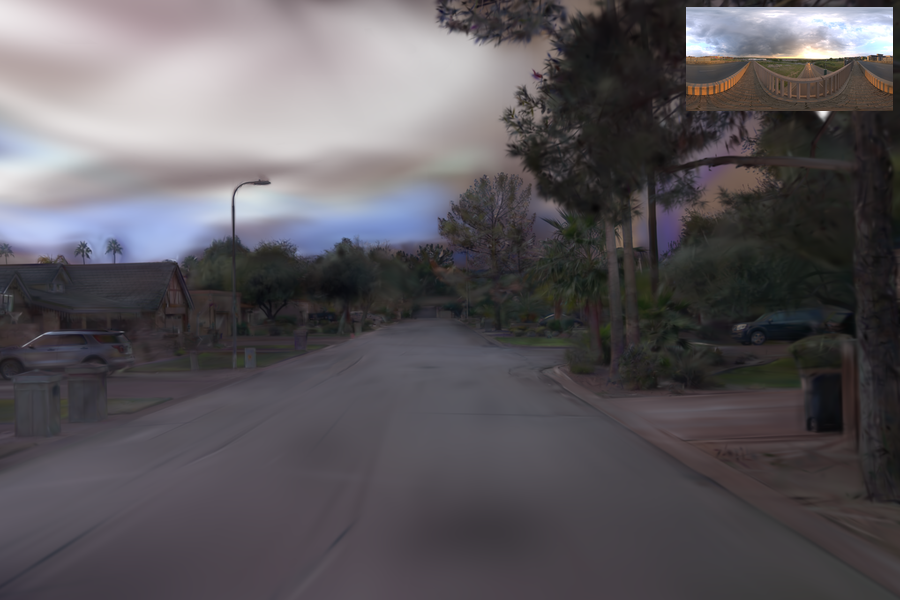} &
\qualcell{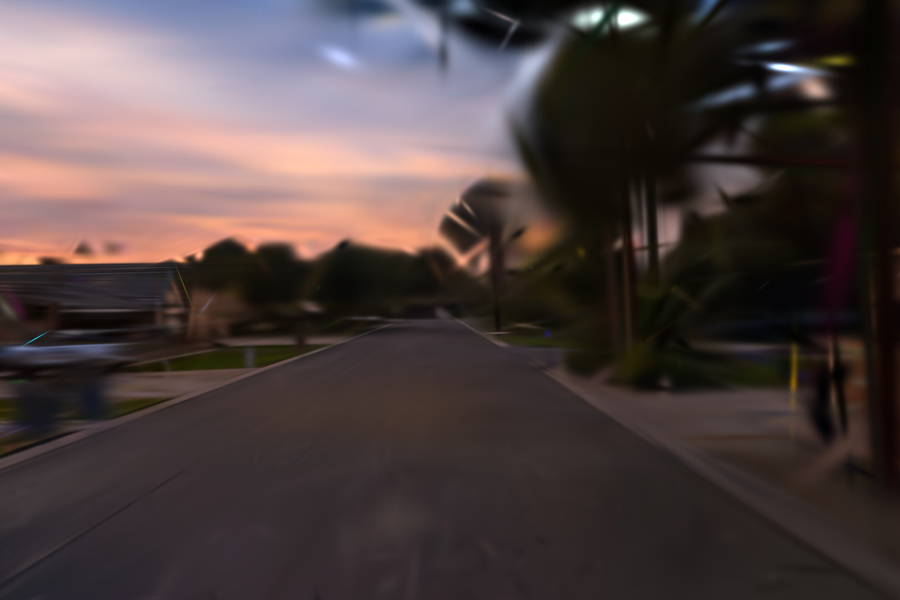} &
\qualcell{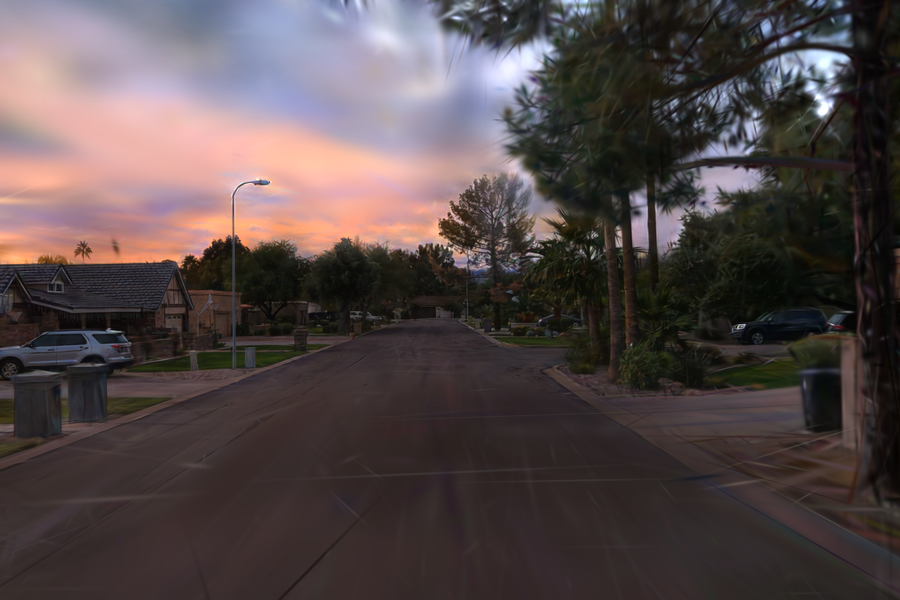} &
\qualcell{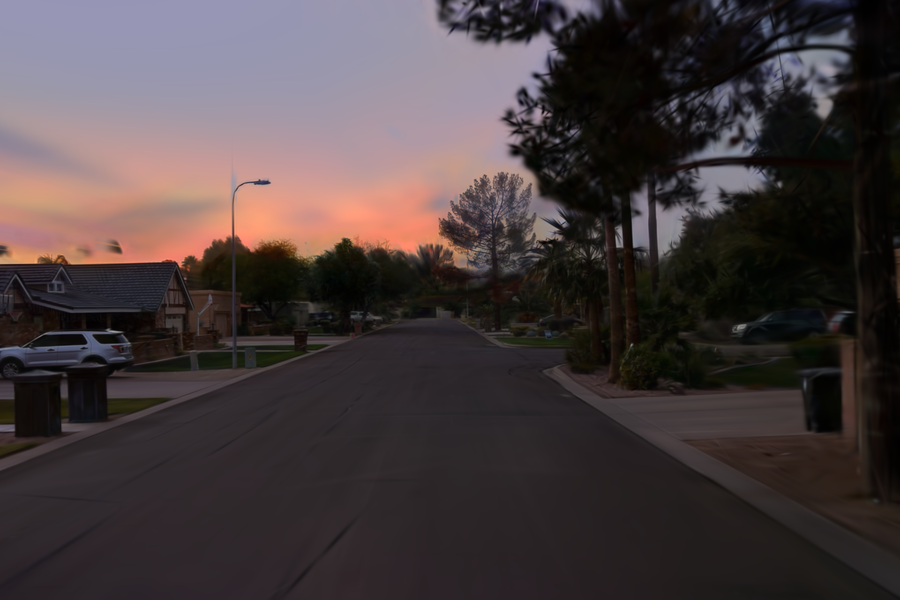}\\[0.3ex]
Blue hour &
\qualcell{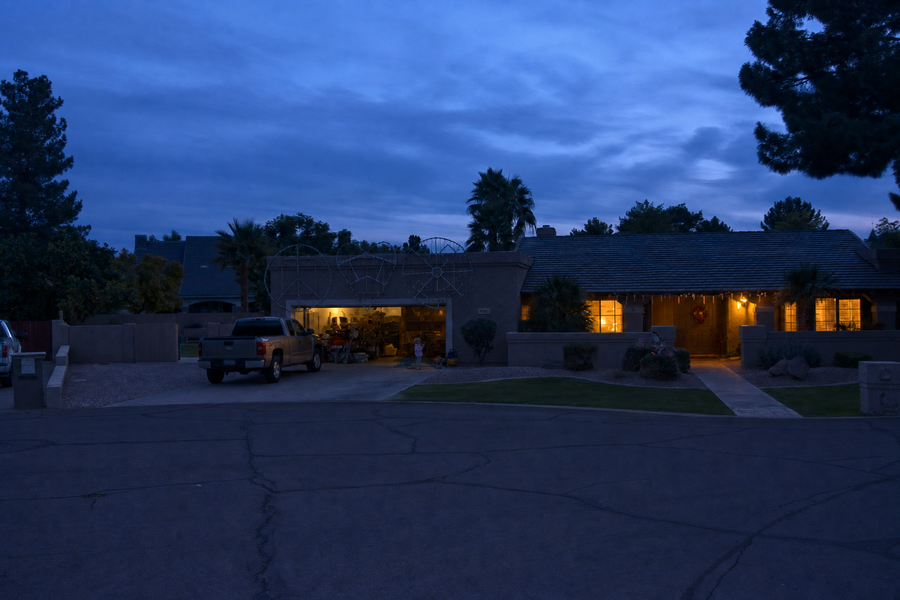} &
\qualcell{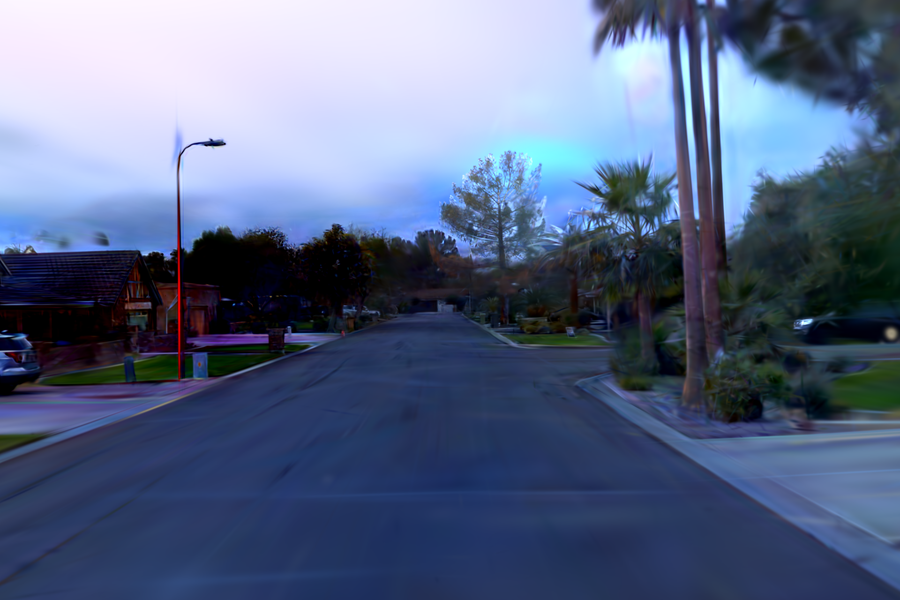} &
\qualcell{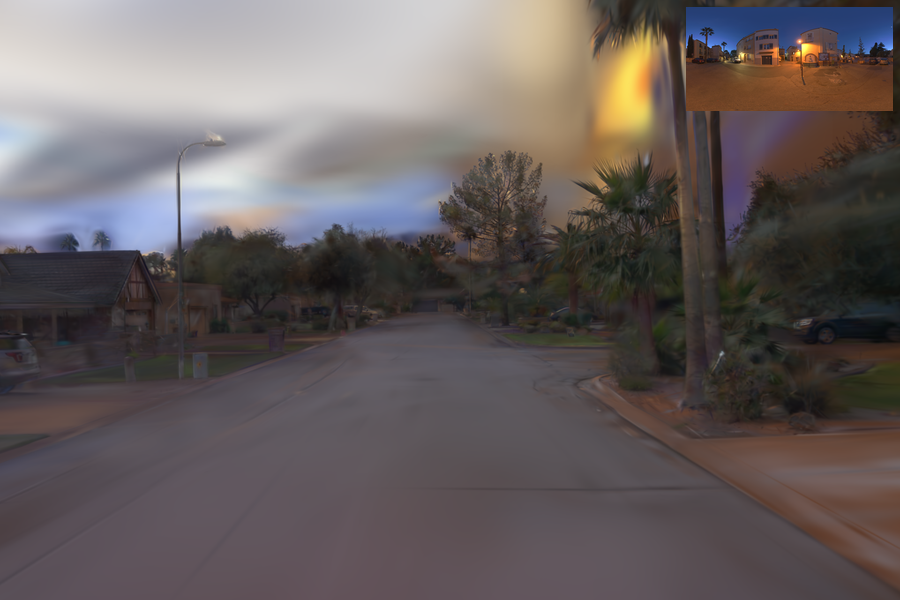} &
\qualcell{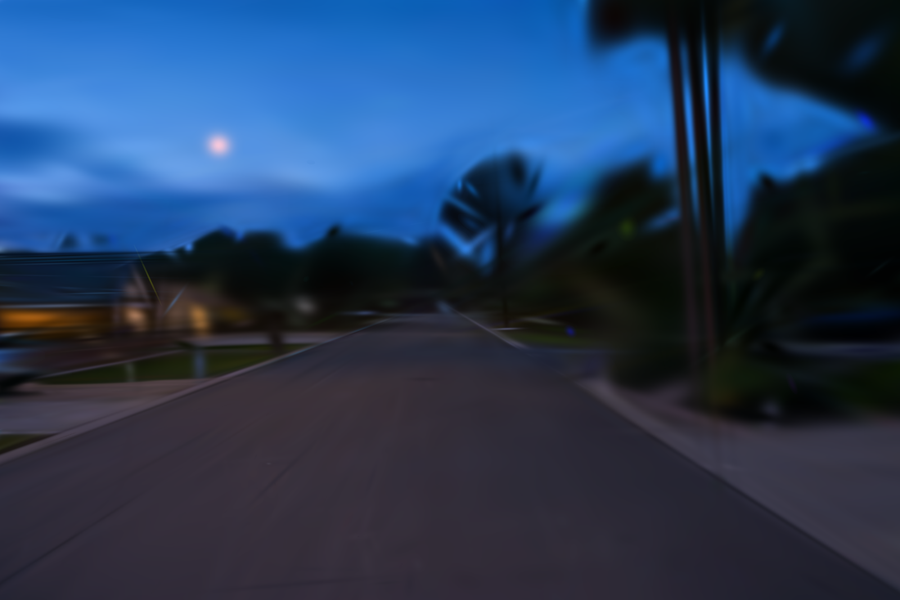} &
\qualcell{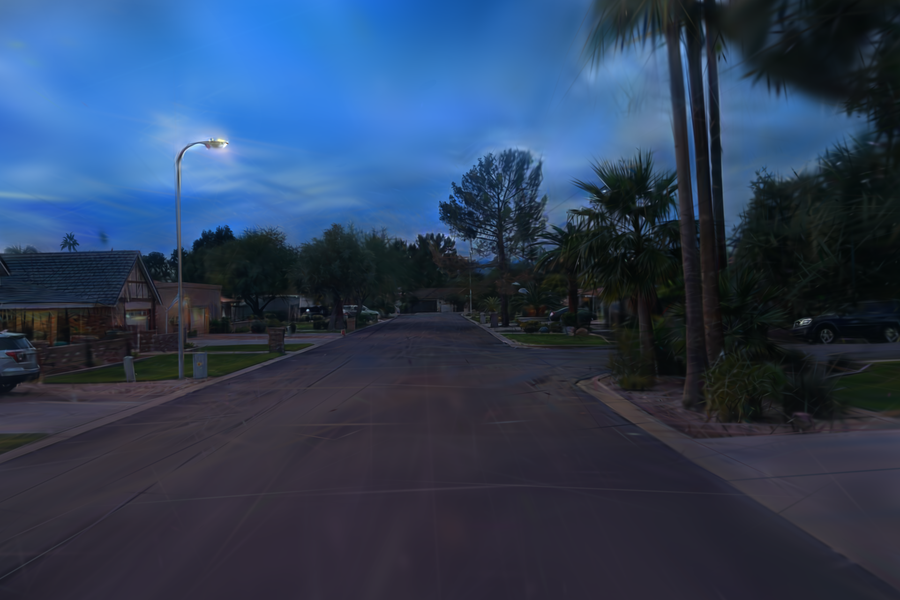} &
\qualcell{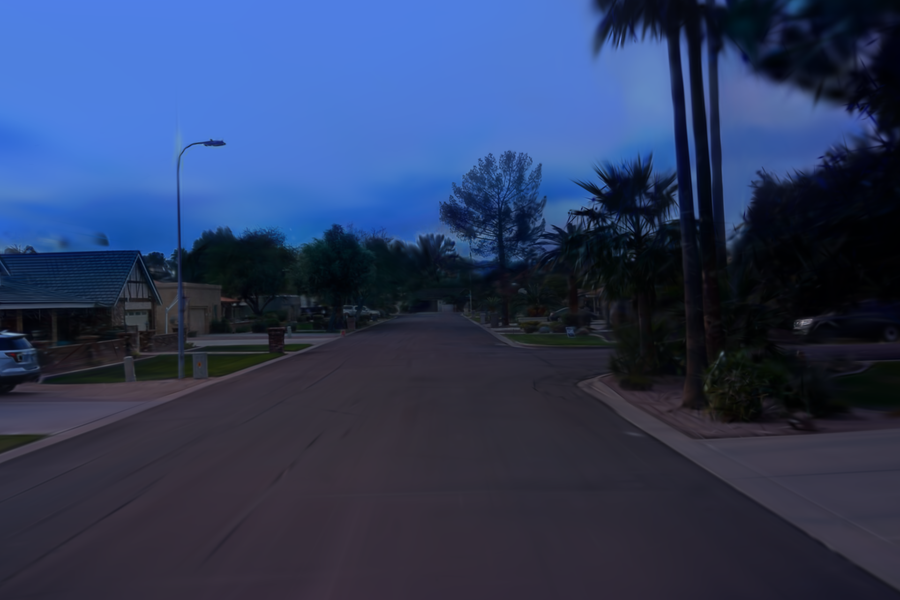}\\[0.3ex]
Overcast &
\qualcell{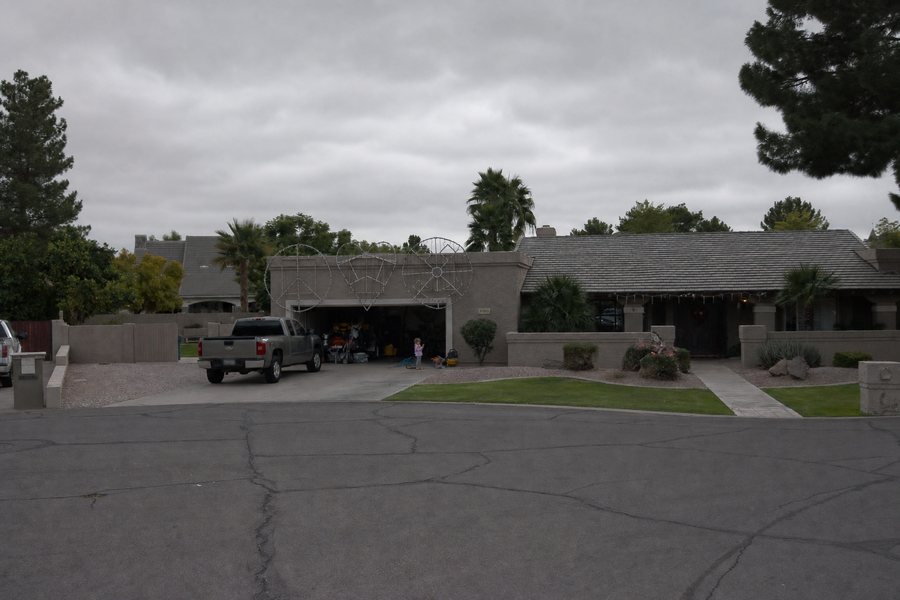} &
\qualcell{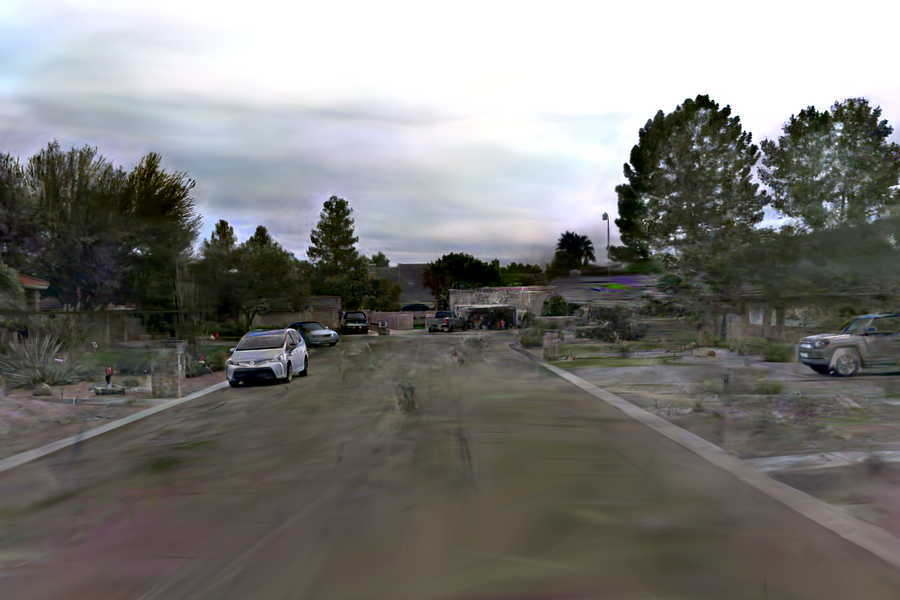} &
\qualcell{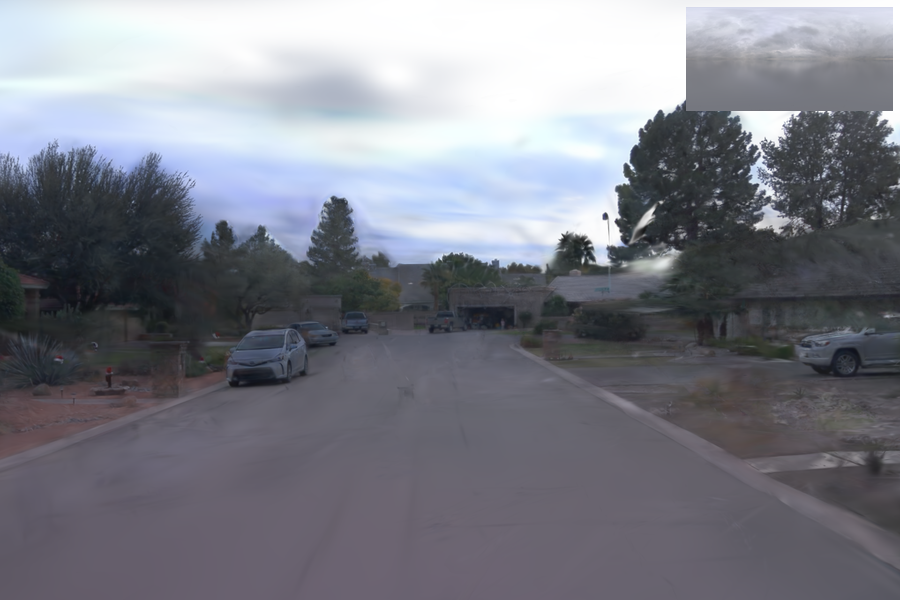} &
\qualcell{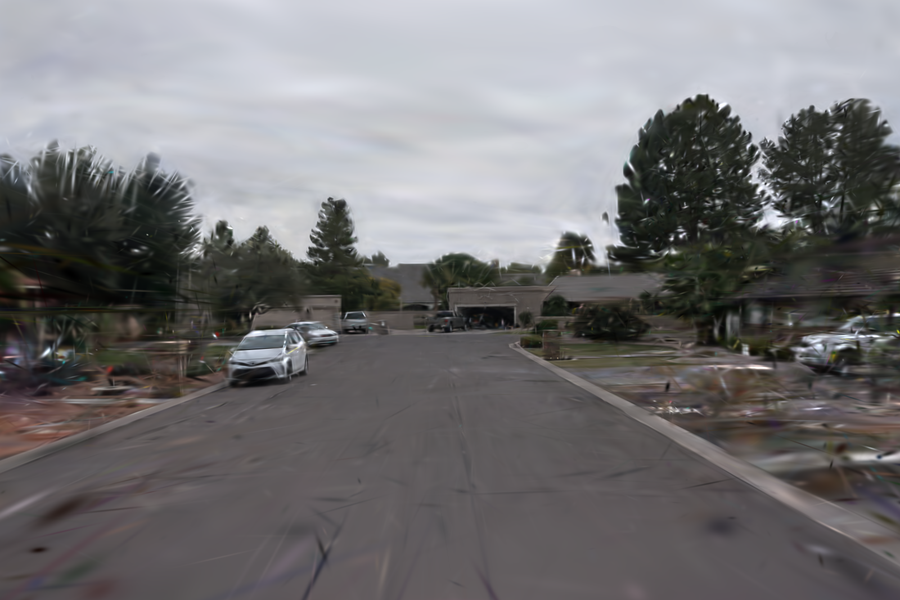} &
\qualcell{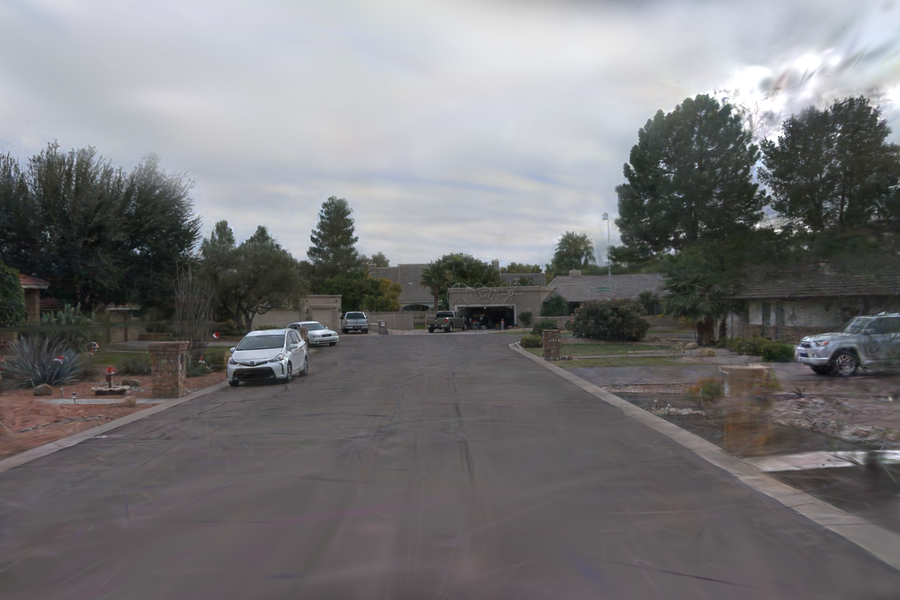} &
\qualcell{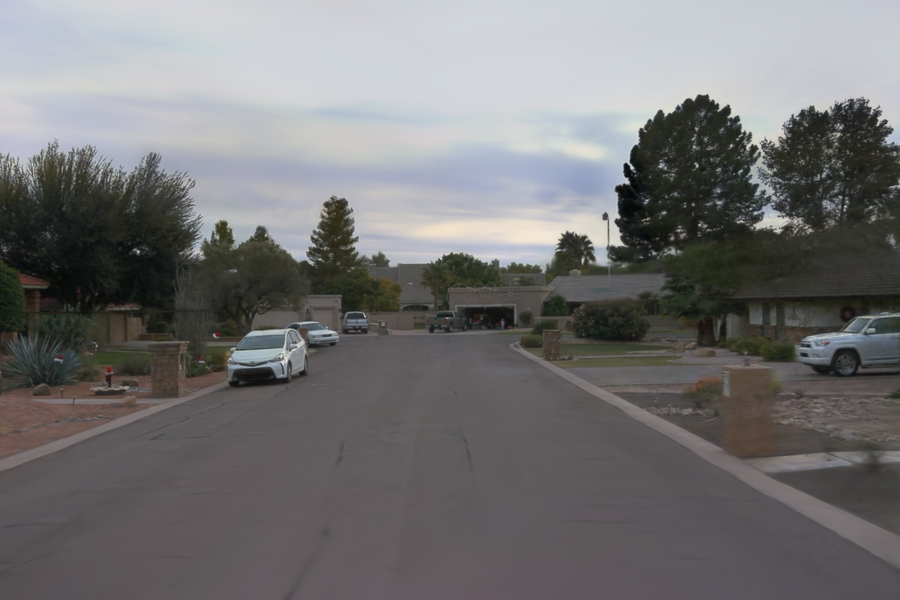}
\end{tabular}
\caption{Qualitative comparison with matched cameras.}
\label{fig:qualitative_main}
\end{figure*}

Fig.~\ref{fig:qualitative_main} compares teacher anchors and five methods on six randomly selected non-anchor matched-camera views from two Waymo scenes. The teacher anchors can contain edit-induced noise on road surfaces and vegetation; artifacts are often retained or amplified by the other baselines such as oversaturation, darkened regions, smoothing, softened boundaries, or locally inconsistent facade and road details. Such editing noise is better suppressed by the proposed method while the requested target appearance is preserved.

Across the three conditions, the proposed method retains recognizable target appearance with fewer artifacts, consistent with Table~\ref{tab:main}.

\subsection{Cross-View Trajectory Consistency}

\begin{figure*}[!b]
\centering
\scriptsize
\setlength{\tabcolsep}{1.2pt}
\renewcommand{\arraystretch}{1.02}
\begin{tabular}{@{}p{0.075\textwidth}ccccc@{}}
\multicolumn{6}{@{}l}{\textbf{Waymo scene / sunset}}\\[-0.2ex]
& \textbf{Frame 135} & \textbf{Frame 136} & \textbf{Frame 137} & \textbf{Frame 138} & \textbf{Frame 139}\\
ViP3DE & \tempcell{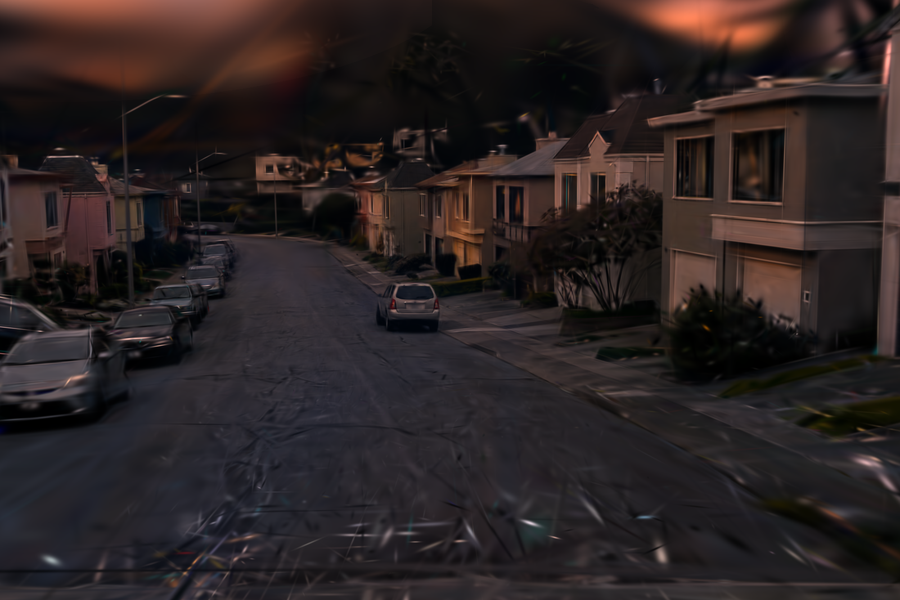} & \tempcell{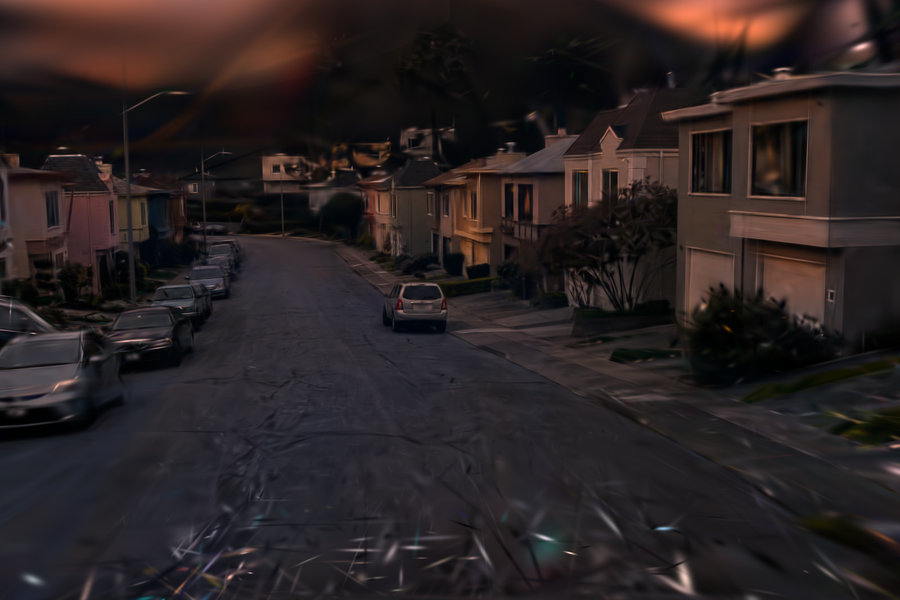} & \tempcell{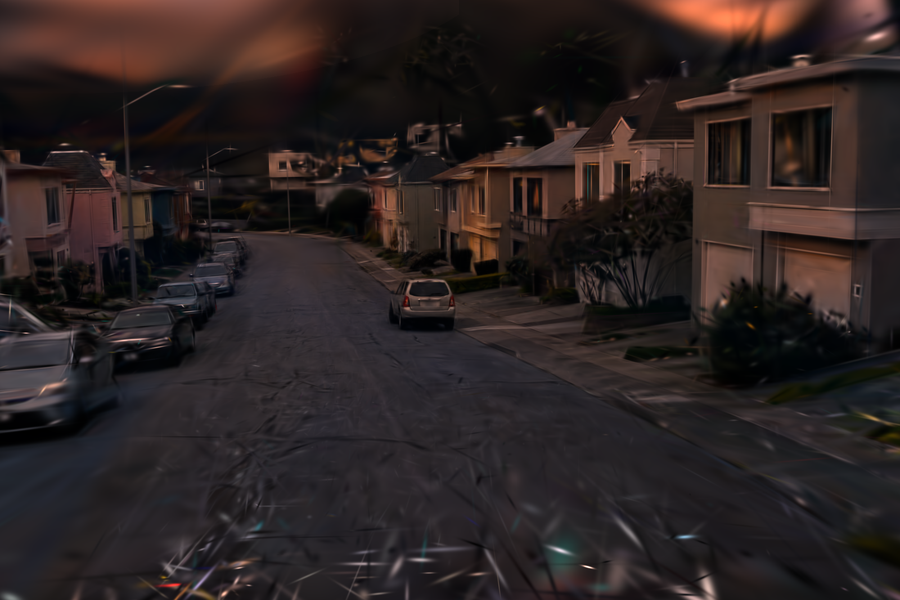} & \tempcell{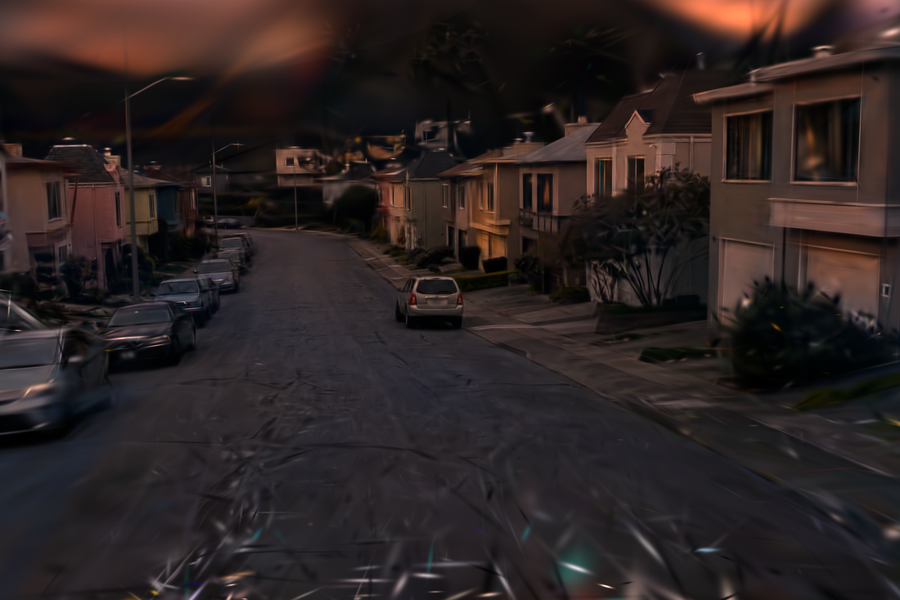} & \tempcell{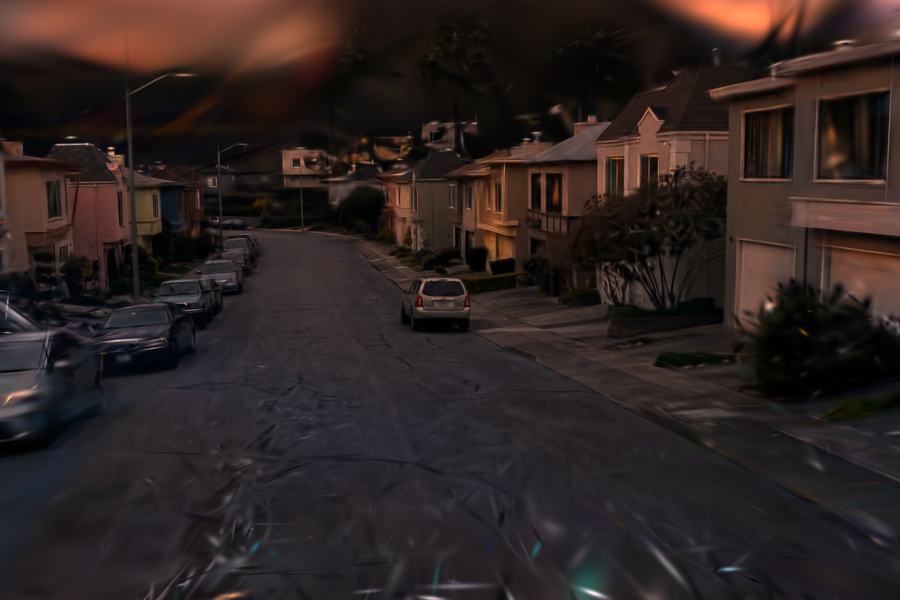}\\[0.3ex]
EditSplat & \tempcell{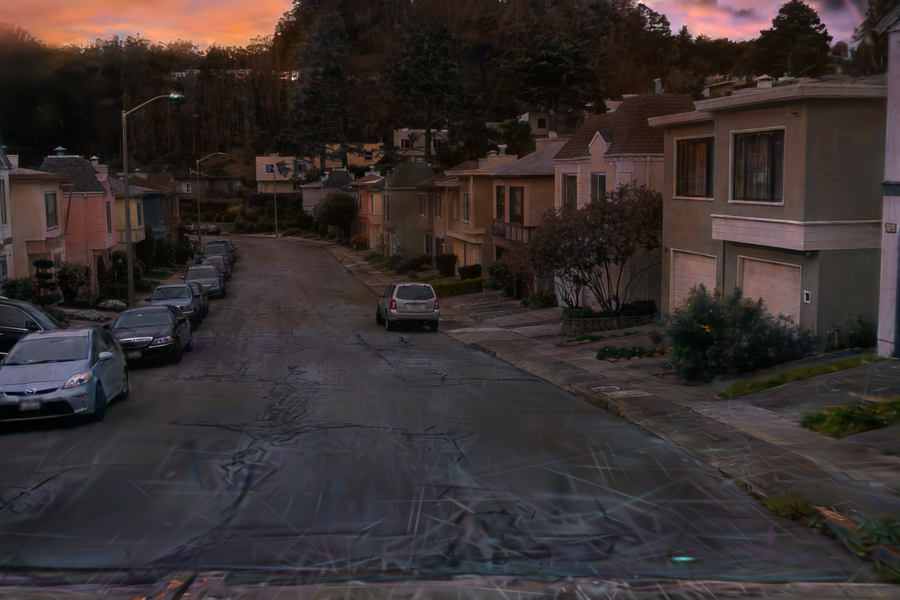} & \tempcell{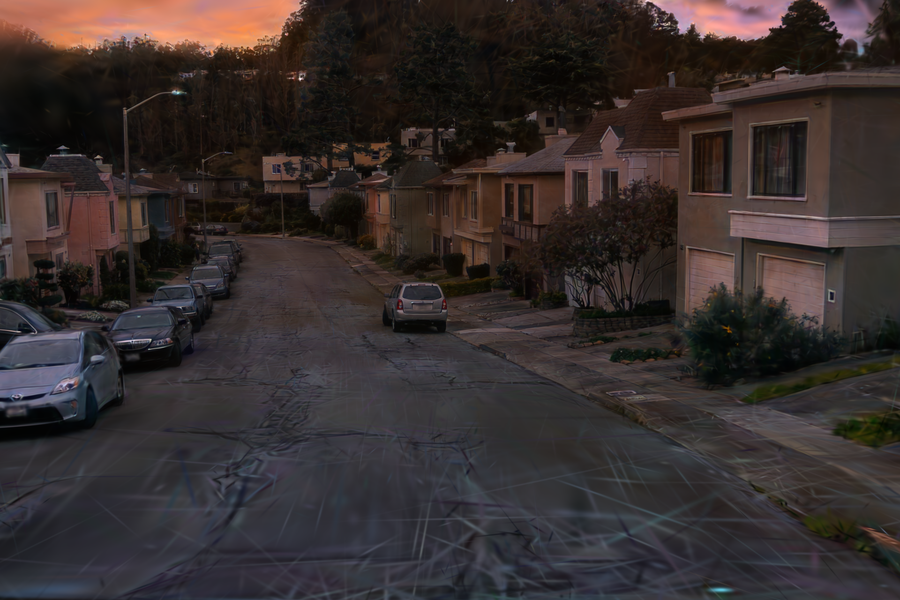} & \tempcell{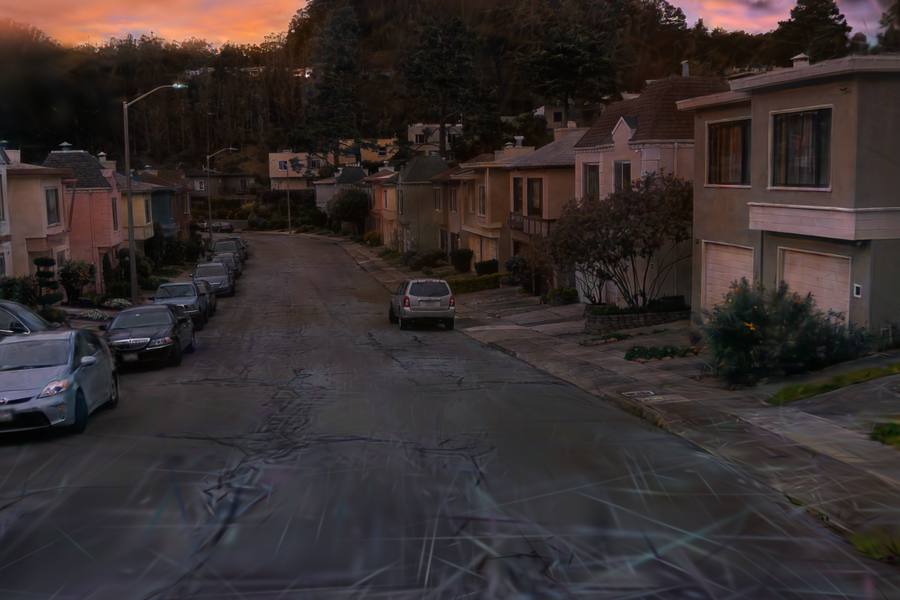} & \tempcell{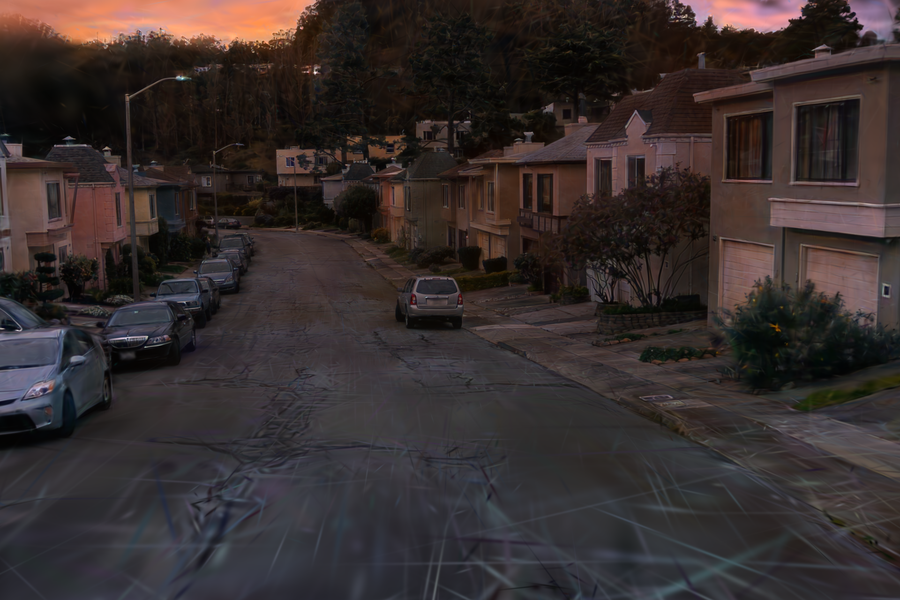} & \tempcell{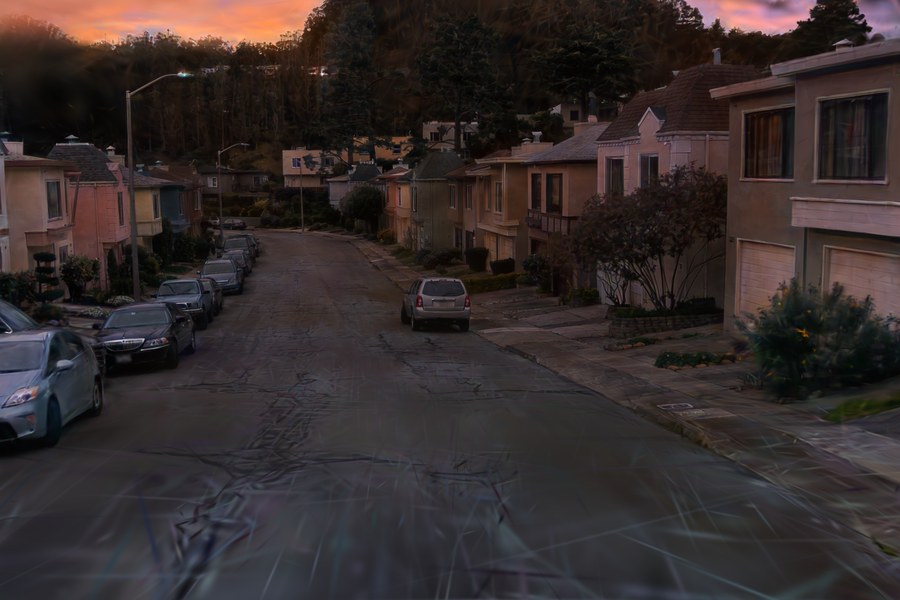}\\[0.3ex]
\textbf{Ours} & \tempcell{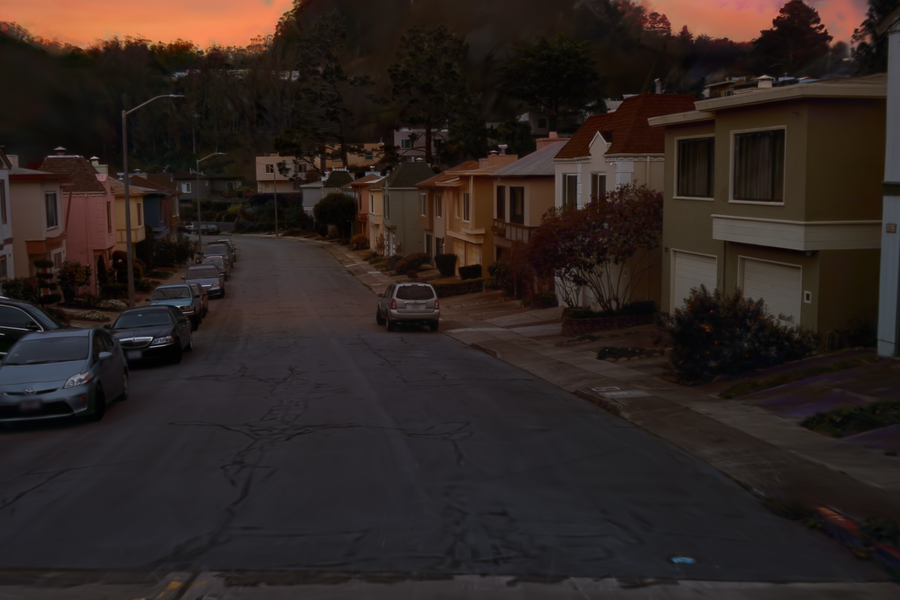} & \tempcell{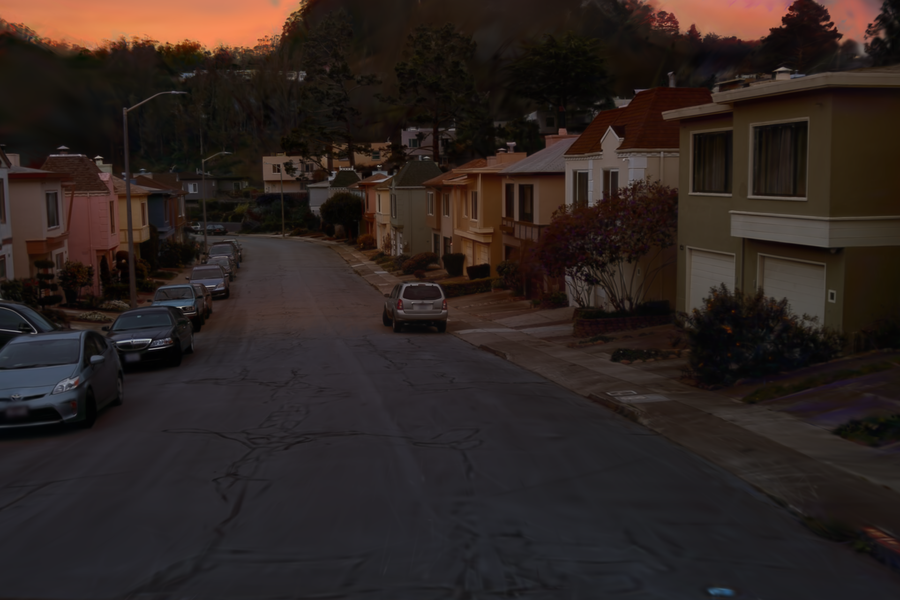} & \tempcell{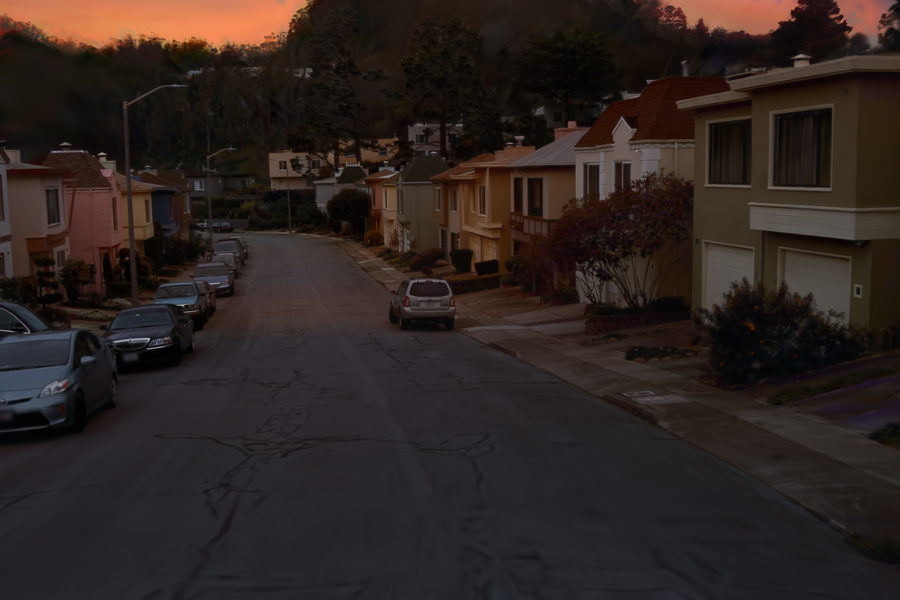} & \tempcell{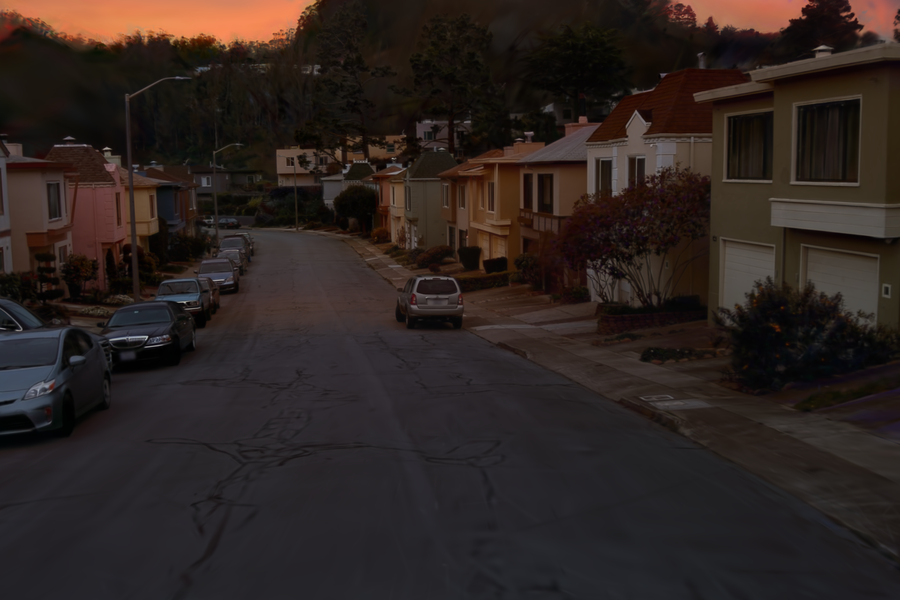} & \tempcell{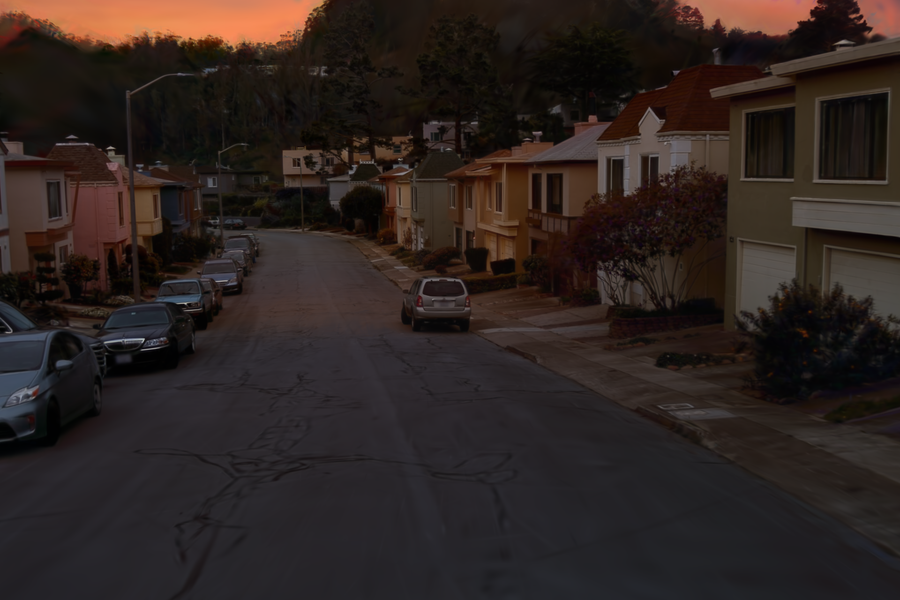}
\end{tabular}
\caption{Five consecutive views of a Waymo scene under sunset.}
\label{fig:temporal_consistency}
\end{figure*}

Fig.~\ref{fig:temporal_consistency} illustrates view-to-view behavior over five consecutive frames of a Waymo sunset sequence. ViP3DE and EditSplat show larger changes in road tone, shadow boundaries, and high-frequency surface patterns, whereas fewer visible editing artifacts and less noise are observed in the proposed sequence, together with a more uniform appearance direction and persistent local structures. This five-frame strip is qualitative only; the CV-$\Delta$ value in Table~\ref{tab:main} uses every valid adjacent pair across all 12 scene--condition groups.

\begin{figure*}[t]
\centering
\scriptsize
\setlength{\tabcolsep}{1.0pt}
\renewcommand{\arraystretch}{1.02}
\begin{tabular}{@{}ccccccc@{}}
\textbf{Teacher} & \textbf{Ours-4} & \textbf{Ours-8} & \textbf{EditSplat-4} & \textbf{EditSplat-8} & \textbf{ViP3DE-4} & \textbf{ViP3DE-8}\\
\includegraphics[width=0.137\textwidth]{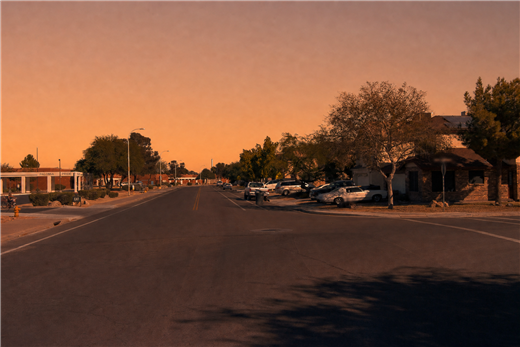} &
\includegraphics[width=0.137\textwidth]{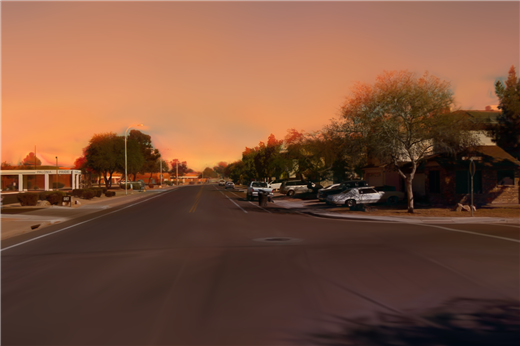} &
\includegraphics[width=0.137\textwidth]{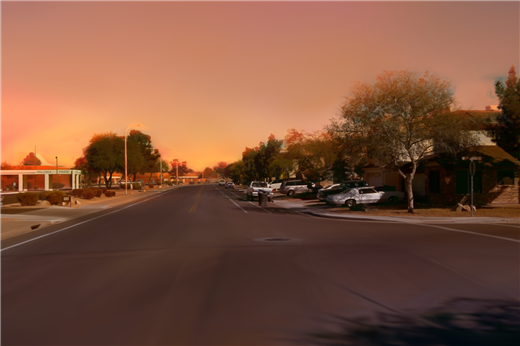} &
\includegraphics[width=0.137\textwidth]{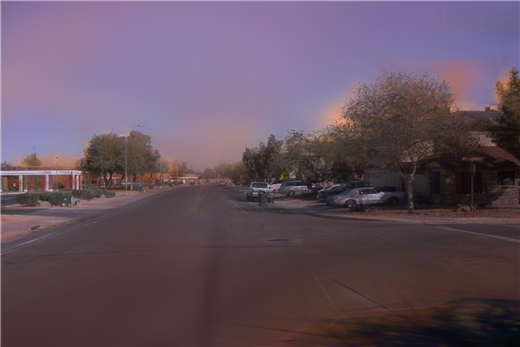} &
\includegraphics[width=0.137\textwidth]{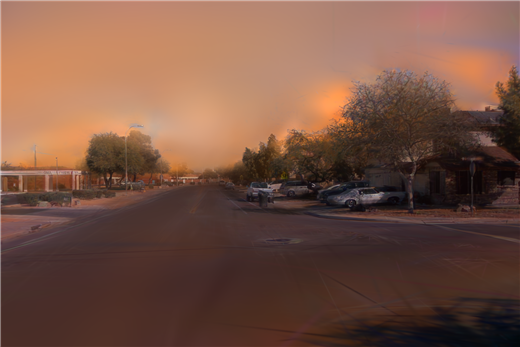} &
\includegraphics[width=0.137\textwidth]{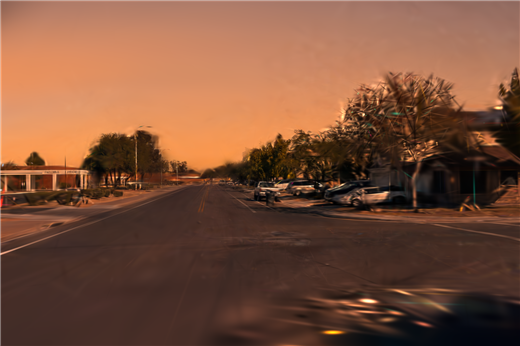} &
\includegraphics[width=0.137\textwidth]{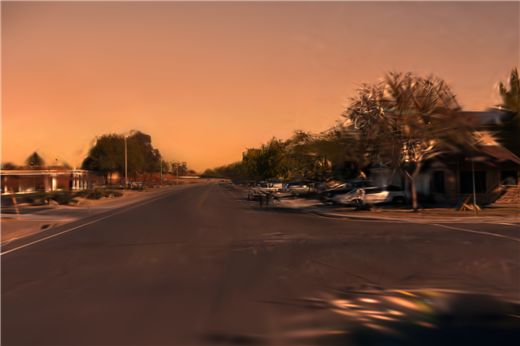}\\[-0.2ex]
\includegraphics[width=0.137\textwidth]{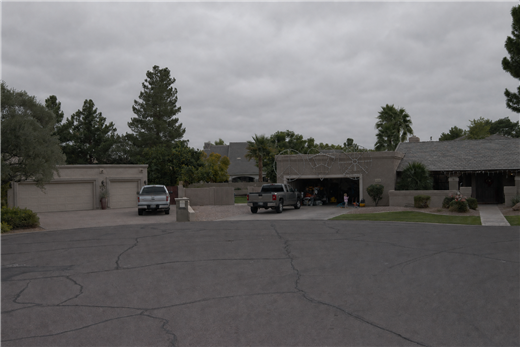} &
\includegraphics[width=0.137\textwidth]{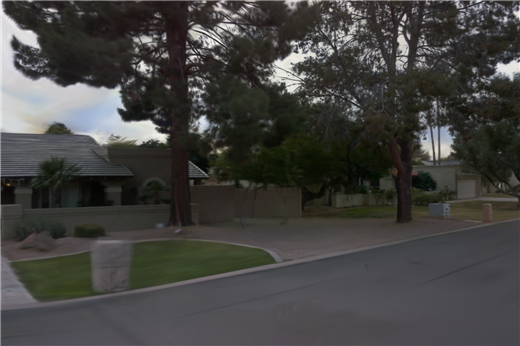} &
\includegraphics[width=0.137\textwidth]{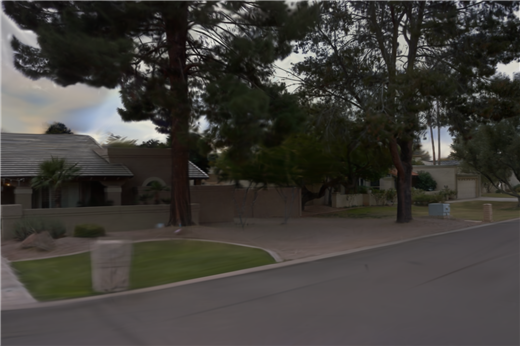} &
\includegraphics[width=0.137\textwidth]{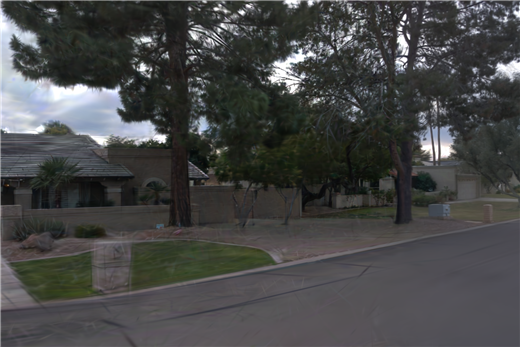} &
\includegraphics[width=0.137\textwidth]{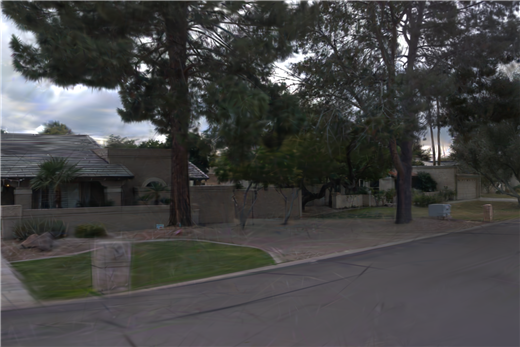} &
\includegraphics[width=0.137\textwidth]{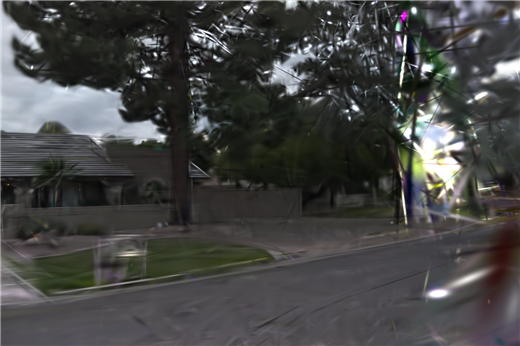} &
\includegraphics[width=0.137\textwidth]{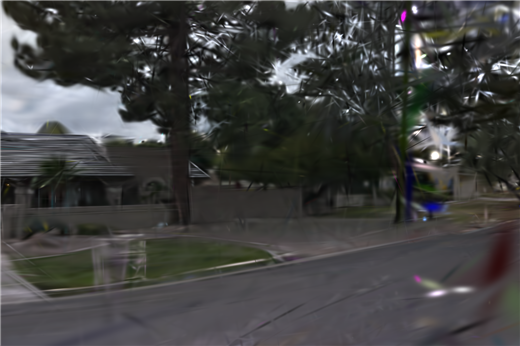}
\end{tabular}
\caption{Sparse-anchor visual comparison with 4 or 8 teacher views.}
\label{fig:sparse_visual_main}
\end{figure*}

\subsection{Ablation Study}

The ablations report target-view fidelity and source-relative non-anchor diagnostics over three representative Waymo scene--condition cases. A-PSNR, A-SSIM, and A-LPIPS compare anchor renderings with teacher targets. N-HP and N-LPIPS measure high-frequency and perceptual deviation from the original Base render on non-anchor views; they are interpreted jointly with target fidelity because smaller deviations can also indicate insufficient editing.

\subsubsection{Teacher-Relative Appearance Residual Distillation}
Table~\ref{tab:ablation_dual_domain} tests the residual-domain design in Fig.~\ref{fig:overview}(b). All variants use the same 30k-iteration Base 3DGS, 16 anchors per case, 182 non-anchor views per case, and 5k appearance-optimization iterations; Renderer-space Only uses the strictly retrained variant. The comparison first contrasts direct sparse teacher fitting with residual-domain baking, then separates the dominant renderer-space signal from the auxiliary Gaussian-space constraint.

\begin{table}[t]
\centering
\caption{Ablation of residual-domain baking; values are three-case averages.}
\label{tab:ablation_dual_domain}
\scriptsize
\setlength{\tabcolsep}{1.4pt}
\begin{tabular}{@{}lccccc@{}}
\toprule
Variant & A-P $\uparrow$ & A-S $\uparrow$ & A-L $\downarrow$ & N-H $\downarrow$ & N-L $\downarrow$\\
\midrule
Absolute Teacher Fitting & \textbf{28.834} & \textbf{0.9812} & \textbf{0.0247} & 0.00442 & 0.27359\\
Gaussian-space Only & 14.837 & 0.7018 & 0.2752 & \textbf{0.00030} & \textbf{0.00568}\\
Renderer-space Only & 23.438 & 0.8077 & 0.3515 & 0.00132 & 0.18005\\
Full Relative-Residual Model & 24.031 & 0.9401 & 0.0586 & 0.00138 & 0.26148\\
\bottomrule
\end{tabular}
\end{table}

Absolute Teacher Fitting gives the strongest anchor reconstruction but larger non-anchor deviation, indicating that sparse RGB fitting propagates stronger changes to unsupervised views. Gaussian-space Only has low N-* values because its anchor-fidelity metrics are substantially worse; this indicates under-editing rather than improved stability.

\begin{figure*}[t]
\centering
\scriptsize
\setlength{\tabcolsep}{1.2pt}
\renewcommand{\arraystretch}{1.02}
\begin{tabular}{@{}ccccc@{}}
\textbf{Base} & \textbf{Absolute Fit} & \textbf{Gaussian Only} & \textbf{Renderer Only} & \textbf{Full}\\
\includegraphics[width=0.175\textwidth]{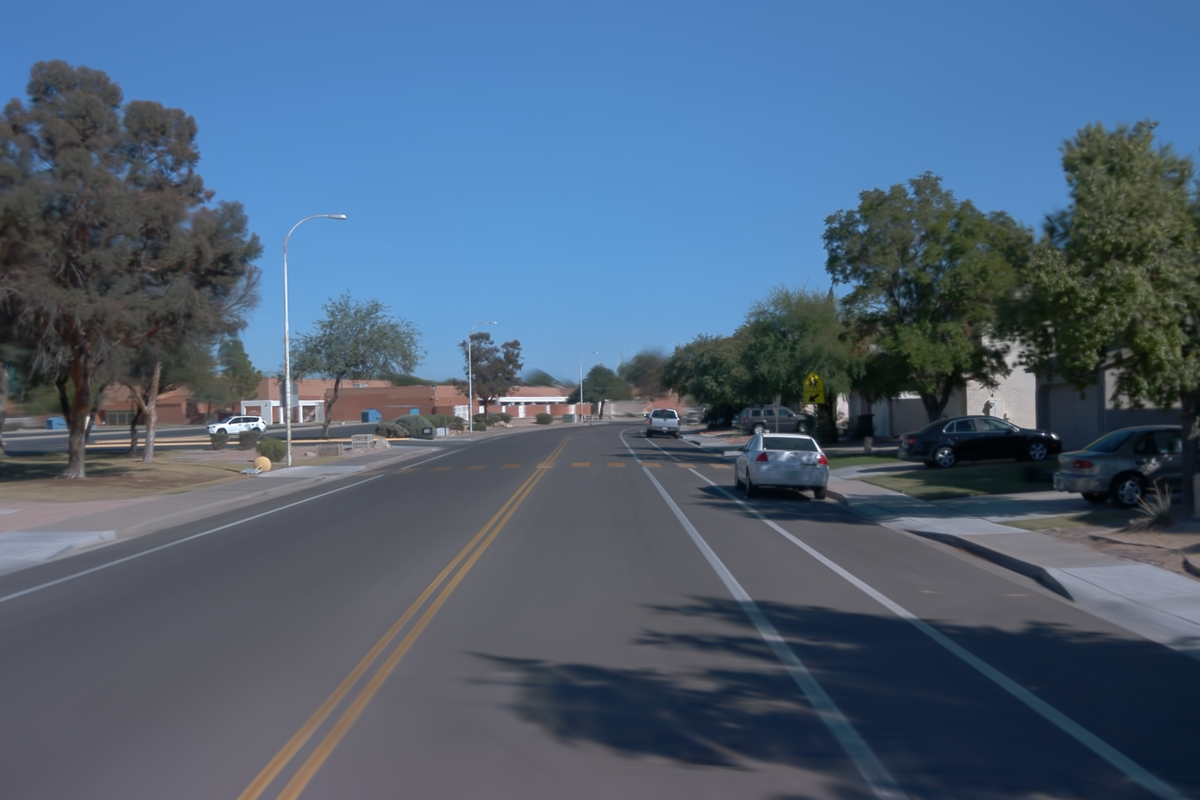} &
\includegraphics[width=0.175\textwidth]{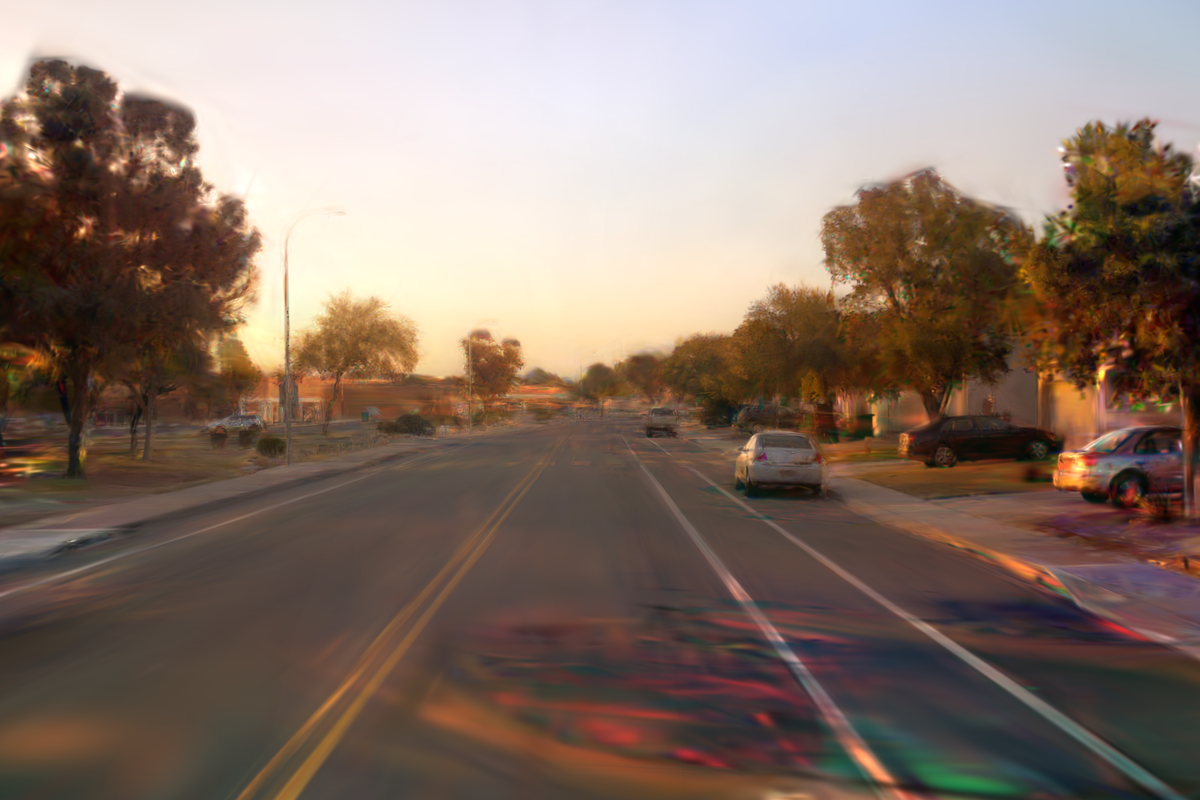} &
\includegraphics[width=0.175\textwidth]{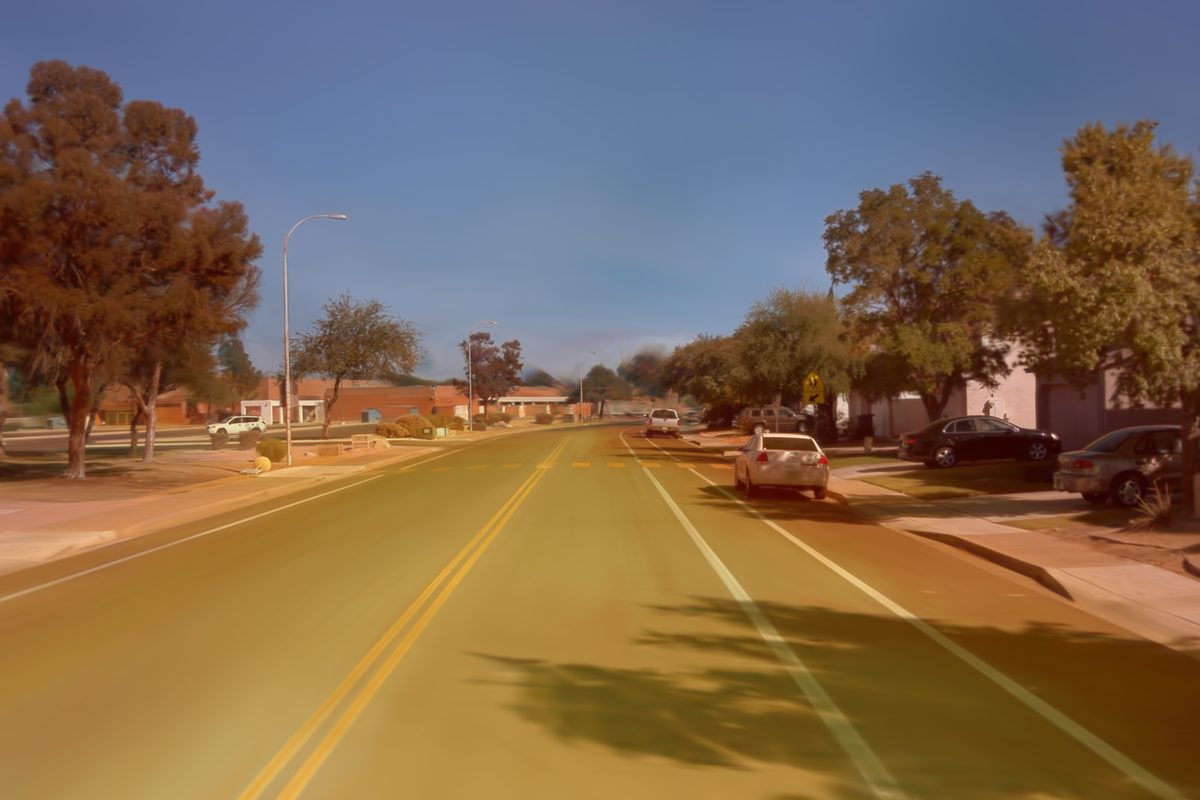} &
\includegraphics[width=0.175\textwidth]{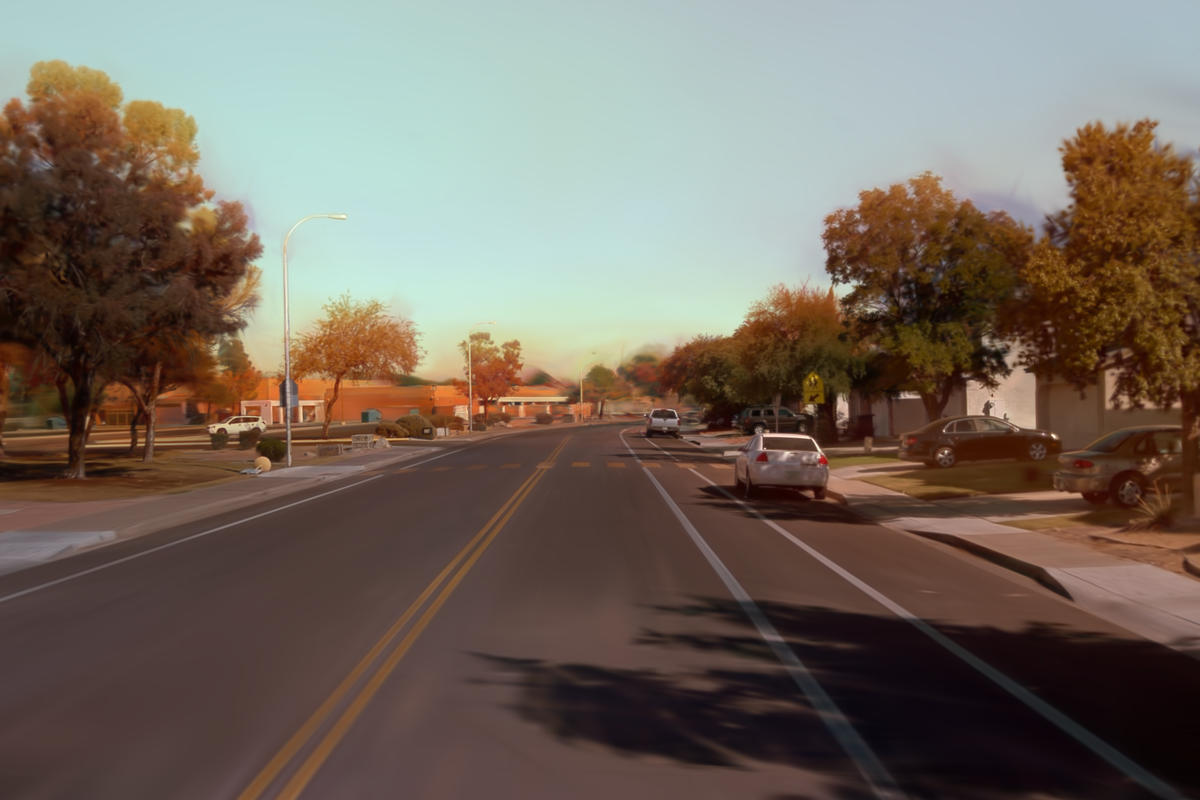} &
\includegraphics[width=0.175\textwidth]{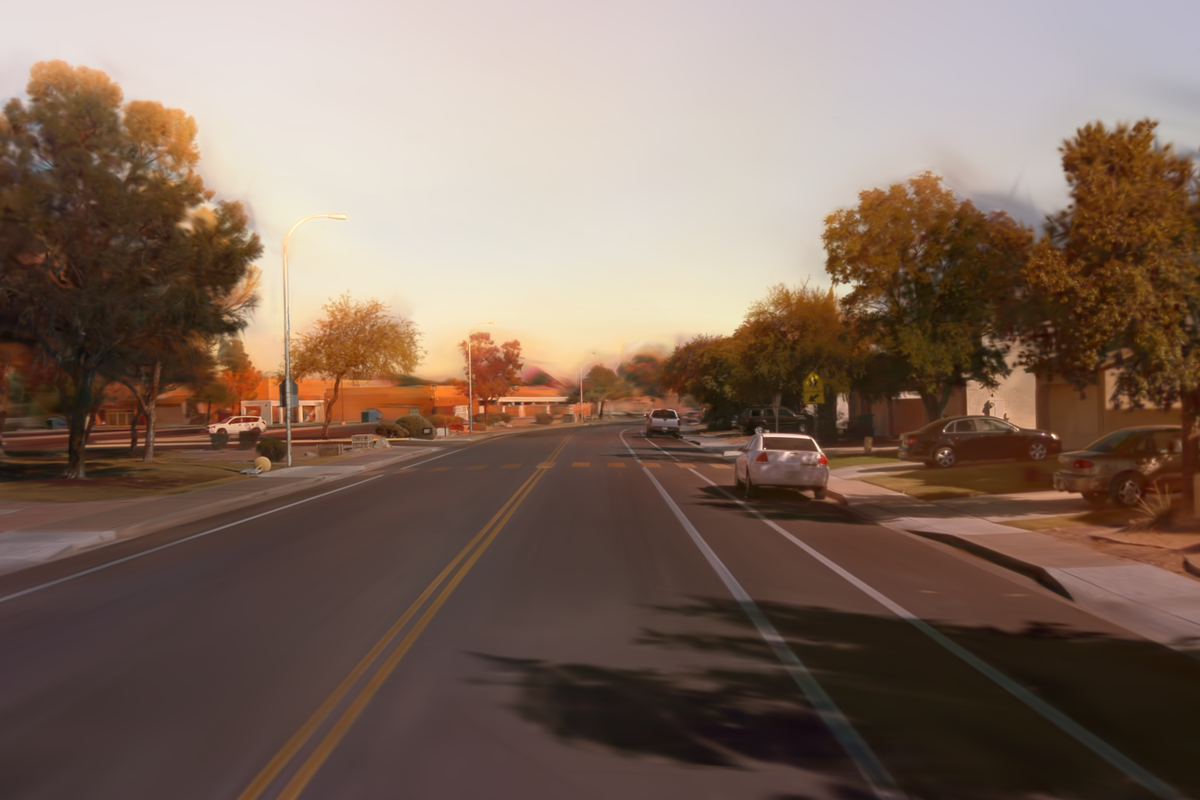}
\end{tabular}
\caption{Residual-domain ablation on a Waymo sunset case using a Qwen image-edit teacher.}
\label{fig:dual_domain_ablation}
\end{figure*}

Renderer-space matching provides the dominant target signal, while Full restores structural and perceptual anchor fidelity through support-aware Gaussian coordination. Fig.~\ref{fig:dual_domain_ablation} shows the same pattern with a Qwen teacher: Absolute Fit contaminates scene structure, Gaussian Only under-edits, and Full preserves cleaner boundaries than Renderer Only.

\subsubsection{Coarse-to-Fine Baked SH Appearance Optimization}
Table~\ref{tab:ablation_coarse_to_fine} aligns the main component removals with Fig.~\ref{fig:overview}(c). Removing the Illumination MLP causes the clearest degradation, reducing A-PSNR by 0.571 dB and A-SSIM by 0.0070 while increasing A-LPIPS by 0.0087. This result supports renderer-space low/mid optimization as the main driver of the global target-condition direction across the three cases.

\begin{table}[t]
\centering
\caption{Component-removal ablation; values are three-case averages.}
\label{tab:ablation_coarse_to_fine}
\scriptsize
\setlength{\tabcolsep}{1.4pt}
\begin{tabular}{@{}lccccc@{}}
\toprule
Variant & A-P $\uparrow$ & A-S $\uparrow$ & A-L $\downarrow$ & N-H $\downarrow$ & N-L $\downarrow$\\
\midrule
Full & \textbf{24.031} & \textbf{0.9401} & \textbf{0.0586} & 0.00138 & 0.26148\\
w/o Illumination MLP & 23.460 & 0.9331 & 0.0673 & 0.00127 & 0.25819\\
w/o Tone Adapter & 23.979 & 0.9394 & 0.0590 & 0.00134 & 0.26081\\
w/o Detail MLP & 23.702 & 0.9358 & 0.0631 & 0.00128 & 0.25868\\
w/o Progressive Auxiliary Decay & 23.852 & 0.9375 & 0.0616 & \textbf{0.00121} & \textbf{0.25699}\\
\bottomrule
\end{tabular}
\end{table}

Full ranks first on all anchor-fidelity metrics, with the Illumination and Detail removals causing the clearest degradation. Lower N-* values in removal variants do not indicate better transfer when anchor fidelity also weakens.

\subsection{Discussion}
Overall, the quantitative and visual evidence shows a favorable balance under sparse-anchor supervision. Across the three anchor budgets, the intended appearance remains recognizable with fewer visible teacher-induced artifacts, and the Train/Truck results show that this behavior extends beyond the Waymo street assets used in the main benchmark. This trend is also supported by the blind human study, which shows consistent preferences across all four perceptual dimensions.

The ablations identify renderer-space matching as the primary signal for target realization. Gaussian-space supervision alone under-edits the target appearance, as reflected by its weaker anchor fidelity, whereas support-aware Gaussian coordination acts as a complementary mechanism that improves structural fidelity and limits unsupported appearance changes in the Full model. The current static-asset setting keeps appearance baking separate from moving-object geometry and occlusion changes; extending the same idea to dynamic street scenes is a natural next step.
\section{Conclusion}

A teacher-conditioned appearance-baking framework is presented for static 3DGS street scenes, combining teacher-relative residuals, confidence-aware lifting, asymmetric detail control, and SH-only baking in a standard-rasterizer 3DGS. Across the evaluated assets and conditions, it provides a favorable overall balance over the baselines, especially by reducing teacher-induced noise in novel views. Ablations show that renderer-space matching is necessary for target realization, while support-aware Gaussian-space coordination improves structural and perceptual anchor fidelity in the full model. Future work will extend the baking strategy to 4D/dynamic Gaussians while preserving fine anchor details.

\FloatBarrier
\IEEEtriggeratref{40}

\end{document}